\documentclass[twoside,11pt]{article}
\usepackage{jair, theapa, rawfonts}

\jairheading{80}{2024}{1187-1222}{06/2023}{07/2024}
\ShortHeadings{A Hybrid Intelligence Method for Argument Mining}
{van der Meer, Liscio, Jonker, Plaat, Vossen \& Murukannaiah}
\firstpageno{1187}

\usepackage{amsmath}
\usepackage{varwidth}
\usepackage[many]{tcolorbox}
\usepackage{xcolor}
\usepackage{calc}
\usepackage[inline]{enumitem}
\usepackage{amssymb}
\usepackage{array}
\usepackage{booktabs}
\usepackage{makecell}
\usepackage{multirow}
\usepackage{multicol}
\usepackage{url}
\usepackage[algo2e, ruled,linesnumbered]{algorithm2e}
\SetKwComment{Comment}{$\triangleright$\ }{}
\usepackage{fontawesome5}
\usepackage{standalone}
\usepackage{longtable}
\usepackage{xspace}
\usepackage{bbm} 
\usepackage{relsize}
\usepackage{pgfplots}
\pgfplotsset{compat=newest}
\usepgfplotslibrary{statistics}

\usetikzlibrary{patterns, calc, shapes, arrows, positioning, matrix, math, external, backgrounds, fit}

\definecolor{bblue}{HTML}{5cb1eb}
\definecolor{rred}{HTML}{be1908}
\definecolor{ggreen}{HTML}{aaad00}
\definecolor{dblue}{HTML}{001158}

\definecolor{paultol1}{HTML}{4477AA}
\definecolor{paultol2}{HTML}{EE6677}
\definecolor{paultol3}{HTML}{228833}
\definecolor{paultol4}{HTML}{CCBB44}
\definecolor{paultol5}{HTML}{66CCEE}
\definecolor{paultol6}{HTML}{AA3377}
\definecolor{paultol7}{HTML}{BBBBBB}

\newtcolorbox[use counter=prompt]{promptbox}[2][]%
   {
    enhanced jigsaw,
    boxrule=0mm,
    colback=white, colframe=gray!50!black,
    top=0mm,bottom=1mm,left=1mm,right=1mm,
    arc=1mm,
    title={\textbf{Prompt \thetcbcounter: #2}},
    borderline={0.5pt}{0pt}{gray!50!black, rounded corners},
    fontupper=\normalsize\sffamily, fontlower=\normalsize\sffamily,
    #1,
    }

\begin{document}

\title{A Hybrid Intelligence Method for Argument Mining}

\author{\name Michiel van der Meer \email m.t.van.der.meer@liacs.leidenuniv.nl \\
       \addr Leiden Institute for Advanced Computer Science (LIACS)\\
       Leiden University
       \AND
       \name Enrico Liscio \email e.liscio@tudelft.nl \\
       \name Catholijn M. Jonker \email c.m.jonker@tudelft.nl \\
       \addr Interactive Intelligence (II)\\
       Delft University of Technology
       \AND
       \name Aske Plaat \email a.plaat@liacs.leidenuniv.nl \\
       \addr Leiden Institute for Advanced Computer Science (LIACS)\\
       Leiden University
       \AND
       \name Piek Vossen \email p.t.j.m.vossen@vu.nl \\
       \addr Computational Linguistics \& Text Mining Lab (CLTL)\\
       Vrije Universiteit Amsterdam
       \AND
       \name Pradeep K. Murukannaiah \email p.k.murukannaiah@tudelft.nl \\
       \addr Interactive Intelligence (II)\\
       Delft University of Technology
}


\newcommand{\hyena}{HyEnA\xspace}
\newcommand{\argkp}{ArgKP\xspace}
\newcommand{\fillnum}{\textcolor{red}{FILL\_N}}
\newcommand{\young}{\textsc{young}\xspace}
\newcommand{\immune}{\textsc{immune}\xspace}
\newcommand{\reopen}{\textsc{reopen}\xspace}

\newcommand{\STAB}[1]{\begin{tabular}{@{}c@{}}#1\end{tabular}}
\newcolumntype{H}{>{\setbox0=\hbox\bgroup}c<{\egroup}@{}}
\newcommand{\ceil}[1]{\left\lceil #1 \right\rceil} 

\maketitle

\begin{abstract}
Large-scale survey tools enable the collection of citizen feedback in opinion corpora. Extracting the key arguments from a large and noisy set of opinions helps in understanding the opinions quickly and accurately. Fully automated methods can extract arguments but (1) require large labeled datasets that induce large annotation costs and (2) work well for known viewpoints, but not for novel points of view. We propose \hyena, a hybrid (human + AI) method for extracting arguments from opinionated texts, combining the speed of automated processing with the understanding and reasoning capabilities of humans. We evaluate \hyena on three citizen feedback corpora. We find that, on the one hand, \hyena achieves higher coverage and precision than a state-of-the-art automated method when compared to a common set of diverse opinions, justifying the need for human insight. On the other hand, \hyena requires less human effort and does not compromise quality compared to (fully manual) expert analysis, demonstrating the benefit of combining human and artificial intelligence.
\end{abstract}

\section{Introduction}
To make decisions on large public issues, such as combating a pandemic and transitioning to green energy, policymakers often turn to the citizens for feedback \shortcite{kythreotis2019citizen,lee2020policy}. This feedback provides insights into public opinion and contains viewpoints from many individuals with different perspectives. Involving the public in the decision-making process helps in gaining their support when the decisions are to be implemented, fostering the legitimacy of the process \shortcite{ostrom1990governing}.

In the face of crises, decisions must be made swiftly. Thus, collecting feedback, analyzing it, and making recommendations ought to be performed under tight time constraints. For example, when deciding on relaxing COVID-19 measures in the Netherlands, researchers had one month to design the experiment, collect public feedback, and make recommendations to the government \shortcite{mouter2021public}. 
The time constraint limits the amount of information researchers can analyze, potentially painting an incomplete picture of the opinions. In the scenario above, researchers processed data manually and they could only analyze less than 8\% of the qualitative feedback provided by more than 25,000 participants. 

Argument Mining (AM) \shortcite{lawrence2020argument} methods can assist in increasing the efficiency of feedback analysis by, e.g., locating and interpreting argumentative feedback and classifying statements as supporting or opposing a decision. However, applying automated AM methods for feedback analysis poses three main challenges. First, AM methods generalize poorly across domains \shortcite{stab2018cross,jakobsen2021spurious,vandermeer2024empirical}. Thus, they require large amounts of domain-specific training data, which is often not available. The use of pretrained language models, with the pre- or fine-tuning paradigm, mitigates but does not solve the reliance on large domain-specific training datasets \shortcite{reimers2019classification,ein2020corpus}.
Second, although AM methods can identify argumentative content, they often do not compress the information \shortcite[e.g.]{chakrabarty2019ampersand,daxenberger2020argumentext}. That is, they struggle to recognize whether two arguments describe the same point of view, leaving the policymakers with the significant manual labor of aggregating arguments \shortcite{korner2021classifying,korner2021castingsentiment}. 
Finally, naively relying on a small sample of labeled data might cause minority opinions to be ignored as they are not well represented \shortcite{klein2012enabling}, creating a bias toward popular (repeated) arguments, which can perpetuate echo chambers and filter bubbles \shortcite{price1989social,schulz2000biased}.

The \emph{key point analysis} (KPA) task \shortcite{bar2020arguments} seeks to automatically compress argumentative discourse into unique \emph{key points}, which can be matched to arguments. However, synthesizing key points is a significant challenge. In the ArgKP dataset, domain experts (skilled debaters) were asked to generate key points. Subsequently, a model was trained to take over the task \shortcite{bar2020quantitative}.
However, the reliance on a few human expert annotators introduces biases of the human experts and may not be representative of the opinions of the larger population. This defeats the purpose of engaging the larger public in a bottom-up deliberative decision-making process.

\begin{figure}[tb]
    \centering
    \includegraphics[width=\textwidth]{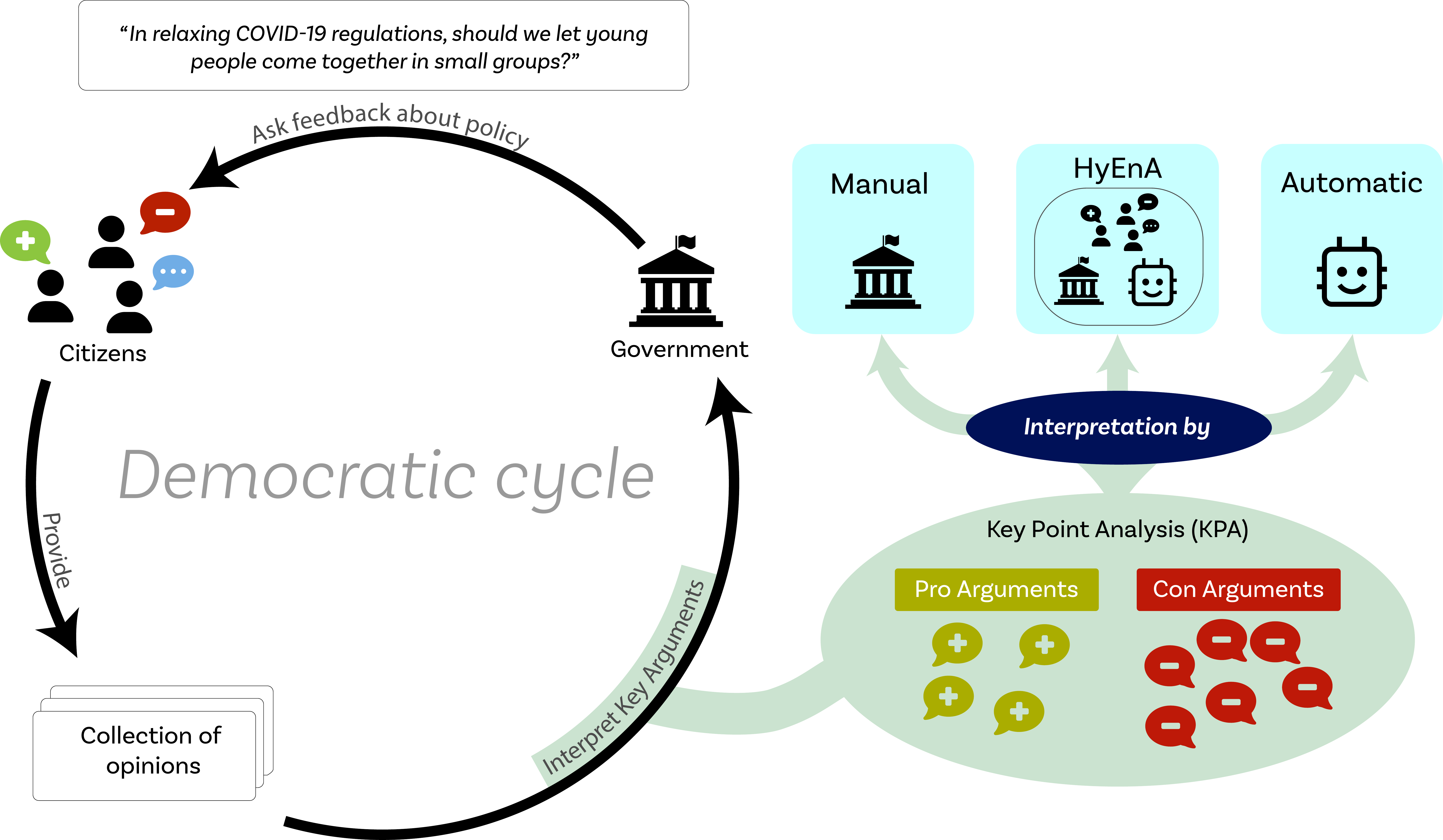}
    \caption{In a democratic cycle, citizens provide their opinions on options for governmental decision-making and their opinions need to be interpreted. Insights into the arguments embedded in their comments can be provided by Key Point Analysis (KPA). To perform KPA, most analysis is performed either manually or automatically. In our work, we propose \hyena, a hybrid method.}
    \label{fig:setup-phase1}
\end{figure}

We argue for a crowd-sourced human-machine approach for argument extraction, combining the scalability of automated methods and the human understanding of others' perspectives. We propose \hyena (\underline{Hy}brid \underline{E}xtractio\underline{n} of \underline{A}rguments), a hybrid (human + AI) method for extracting a diverse set of key arguments from a textual opinion corpus. \hyena breaks down the argument extraction task into argument \emph{annotation}, \emph{consolidation}, and \emph{selection} phases. \hyena employs human (crowd) annotators and supports them via intelligent algorithms based on natural language processing (NLP) techniques for analyzing opinions provided by a large audience, as shown in Figure~\ref{fig:setup-phase1}.

\hyena is evaluated on three corpora, each containing more than 10K public opinions on relaxing COVID-19 restrictions \shortcite{mouter2021public}.  We compare \hyena with an automated approach \shortcite{bar2020quantitative} performing the KPA task. In addition, we compare the key arguments generated by \hyena with manually obtained insights identified by experts \shortcite{mouter2021public}. We find that \hyena outperforms the automated baseline in terms of precision and diversity, specifically when confronted with a set of varied perspectives. \hyena also yields better results than manual analysis, as fewer opinions needed to be analyzed in order to obtain a wider set of key arguments. 

\paragraph{Contributions}
\begin{description}
    \item [C1] We present a hybrid method for key argument extraction, which generates a diverse set of key arguments from a collection of opinionated user comments.
    \item [C2] We evaluate our method on real-world corpora of public feedback on policy options. Compared to an automated baseline, \hyena increases the precision of the key arguments produced and improves coverage over diverse opinions. Compared to the manual baseline, \hyena identifies a large portion of arguments identified by experts as well as new arguments that experts did not identify.
    \item [C3] We extensively discuss the implications of incorporating recent advances in NLP, such as Large Language Models (LLMs), into the workflow of our hybrid method. 
\end{description}

\paragraph{Extension}
In this paper, we extend the \hyena method \shortcite{vandermeer2022hyena} to include \emph{argument selection}. The original \hyena method outputs argument clusters, and leverages manual annotations from the first two phases to select arguments from argument clusters. In this extension, we introduce a method for selecting the most representative argument from each cluster. The need to summarize argument clusters is not specific to \hyena, as previous AM applications also retrieve clusters instead of singular arguments \shortcite{boltuvzic2015identifying,wachsmuth2018retrieval,daxenberger2020argumentext}. We compare various techniques to accomplish this task, including generative large language models. Furthermore, we run additional experiments to demonstrate how the new argument selection step can be incorporated into the \hyena pipeline, and rerun the original evaluation to compare between \hyena with and without the inclusion of argument selection. Finally, we perform additional analyses to derive further insights from annotators in \hyena. We also provide our code, annotation guidelines, and experimental details in the supplementary materials \shortcite{vandermeer2024hyenasuppl}.

\paragraph{Organization} Section~\ref{sec:related-work} provides background on Argument Mining for public opinions, and Section~\ref{sec:method} introduces the \hyena method for extracting arguments. We outline the experimental setup in Section~\ref{sec:experimental} and provide extensive results in Section~\ref{sec:results}. A discussion of our work is given in Section~\ref{sec:discussion} and we conclude with Section~\ref{sec:conclusion}.

\section{Related work}
\label{sec:related-work}
We describe related work on Argument Mining, methods for summarizing arguments, and their application to opinion analysis.

\subsection{Computational Argument Analysis}
\label{sec:am}
Argument Mining (AM) methods \shortcite{cabrio2018five,lawrence2020argument} focus on the recognition, extraction, and computational analysis of arguments presented in natural language. They seek to discover arguments brought forward by speakers and identify connections between them. Typically, AM techniques concern themselves with finding the \emph{structure} of arguments \shortcite{eemeren1987handbook}, with the goal of finding premises for supporting or refuting conclusions.

AM is a challenging problem. The ability to recognize and extract arguments from text (for humans and machines, alike) is dependent on the argumentativeness of the underlying data. Often, significant effort is required by human annotators to reach moderate inter-rater agreement when annotating arguments \shortcite{teruel2018increasing}. Given argumentative texts, modern NLP models are reasonably good at recognizing argumentative discourse within specific contexts \shortcite{niculae2017argument,eger2017neural,reimers2019classification}.

Typically, the first step of AM is to identify the elemental components of arguments (e.g., \emph{claims} and \emph{premises}) in text \shortcite{palau2009argumentation}. The combination of such components forms a structured argument. However, there is currently no consensus on the exact linguistic notion of such elemental components, with multiple levels of granularity being proposed \shortcite{daxenberger2017essence,walton2008argumentation,freeman2011argument,bentahar2010taxonomy}. Nonetheless, a few characteristics have been recognized as important for recognizing arguments, namely that arguments 
\begin{enumerate*}[label=(\arabic*)]
    \item contain (informal) logical reasoning \shortcite{stab2014annotating},
    \item address a \emph{why} question \shortcite{biran2011identifying}, and
    \item have a non-neutral stance towards the issue being discussed \shortcite{stab2014annotating}.
\end{enumerate*}

\hyena is a novel AM method that combines human annotators and automated NLP models. By splitting up the argument extraction task into distinct phases, we take advantage of the diverse human perspectives, while addressing scalability through automation. 

\subsection{Summarization of Arguments}
Automated methods have been proposed to derive high-level insights from large-scale argumentative content. For instance, these approaches focus on indexing and searching through arguments \shortcite{stab2018argumentext,wyner2012semi}, or creating visual overviews of argument structures \shortcite{khartabil2021design,caillou2020cartolabe}. While these may provide access to argumentative content, they are limited in providing a single high-level overview of the arguments on a topic of discussion. Instead, we turn our focus to approaches that create a comprehensible text-based summary from a large corpus of individual comments \shortcite[e.g.]{bar2020quantitative}. In this paradigm, comments are filtered by a manually tuned selection heuristic, resulting in a list of key point candidates. The candidates are matched against all comments, based on a classifier trained for the argument--key point matching task \shortcite{bar2020arguments}. Such approaches have been applied in multiple domains, showcasing their applicability across context \shortcite{barhaim2021every} at varying levels of granularity \shortcite{cattan2023key}. While these approaches present high-level arguments, they struggle to capture diversity in opinions, which is important for accommodating multiple perspectives \shortcite{vandermeer2024empirical}. In this work, we evaluate the performance of these approaches on a novel domain of COVID-19 measures and compare it against \hyena.

Additionally, there exists an extended body of work on Natural Language Inference (NLI) and Semantic Textual Similarity (STS). In these works, models are trained to indicate semantic similarity or logical entailment between two sentences \shortcite{conneau2017supervised,reimers2019sentence}. They have made a significant impact across a range of tasks \shortcite{xu2018deep,zhong2020extractive}. However, downstream applications often need additional fine-tuning \shortcite{howard2018universal} in order to perform a task well. They also capture generic aspects of semantic similarity and entailment, which may not be applicable to arguments \shortcite{reimers2019sentence}, or overfit to spurious patterns in the data \shortcite{mccoy2019right}. Thus, such methods require significant adaptation to effectively compress information in particular domains. Recently, Large Language Models (LLMs) have been shown to perform well on inference tasks with out-of-distribution data \shortcite{wang2023robustness}. However, we argue that a plurality of (human) perspectives is necessary to perform sensitive tasks such as the summarization of arguments, which may in turn be used to inform policy-makers about the sentiment of a population \shortcite{Talat2022OnLanguage}. Yet, LLMs might be adequate for specific subtasks, as we showcase in the third phase of the \hyena method.

\section{Method}
\label{sec:method}
\hyena is a hybrid method since it combines automated techniques and human judgment \shortcite{akata2020research,DellAnna-2024-AAMAS-HIQuality}. \hyena guides human annotators in synthesizing \textit{key arguments} (i.e., high-level semantically distinct arguments that describe relevant aspects of the topic under discussion) from an \textit{opinion corpus} composed of individual \textit{opinions} (textual comments) on a topic. Key arguments are high-level and summarize a group of arguments, similar to key points as introduced by \shortcite{bar2020arguments}. We adopt the term key argument, to emphasize their argumentative nature, as opposed to more generic extractive summarization \shortcite[e.g.]{see2017get}.

\begin{figure}[tb]
    \centering
    \includegraphics[width=\textwidth]{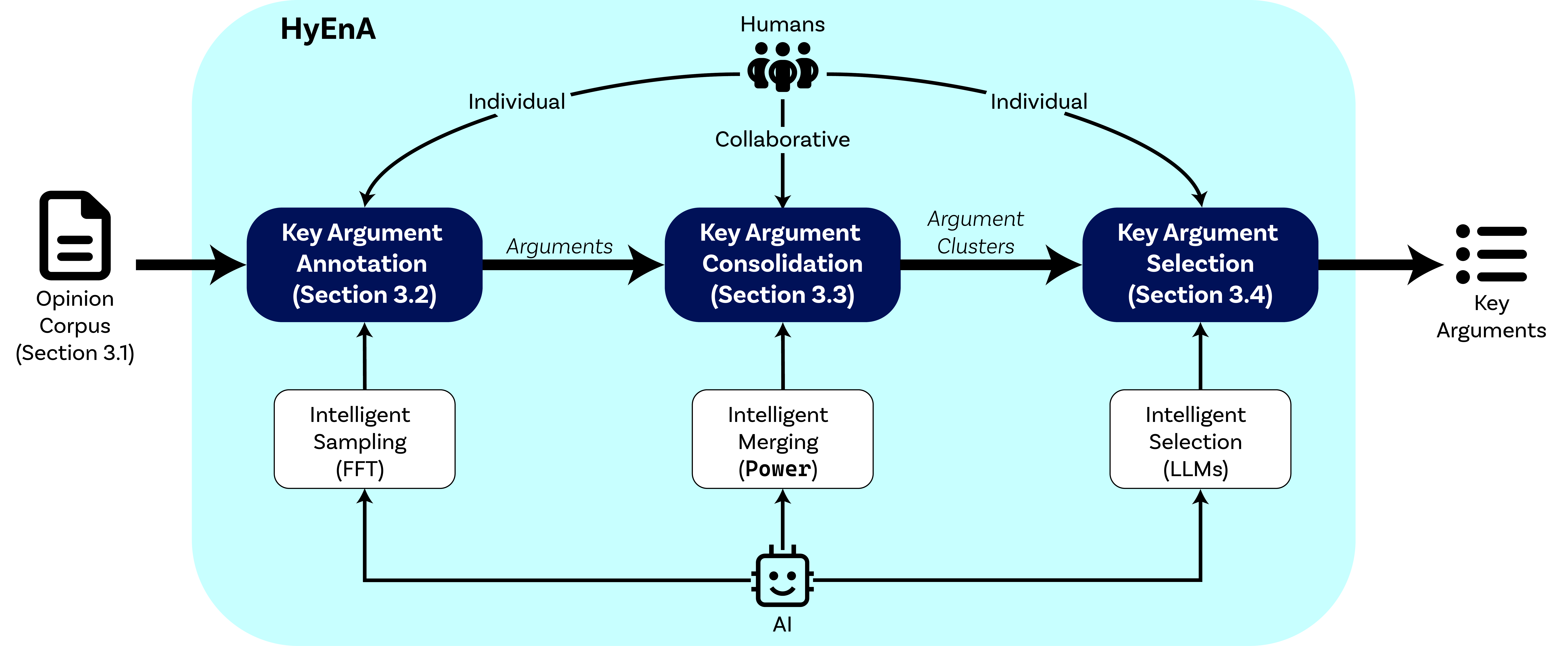}
    \caption{Overview of the \hyena method.}
    \label{fig:hyena-overview}
\end{figure}

\hyena consists of three phases (Figure~\ref{fig:hyena-overview}). In the first phase (\emph{Key Argument Annotation}), an intelligent sampling algorithm guides human annotators individually through an opinion corpus to extract high-level information from the opinions.
In the second and third phases, \hyena aims to reduce the subjectivity in the first phase annotations by combining and rewriting arguments that were individually annotated.
In the second phase (\emph{Key Argument Consolidation}), an intelligent merging strategy supports a new group of annotators in merging the results from the first phase into clusters of arguments, combining manual and automatic labeling. In the third phase (\emph{Key Argument Selection}), \hyena employs an automated method to synthesize a single argument that represents the arguments belonging to the same merged argument cluster. The final output of \hyena is a list of key arguments grounded on the opinions in the corpus. 

\subsection{Opinion Corpora}
\label{sec:PVE}
Our opinion corpora are composed of citizens' feedback on COVID-19 relaxation measures, a contemporary topic. 
The feedback was gathered in April and May 2020 using the Participatory Value Evaluation (PVE) method \shortcite{mouter2021public}. In a PVE, participants are offered a set of policy options and asked to select their preferred portfolio of choices. Then, the participants are asked to explain why they picked certain options (\emph{pro} stance) and not pick the other options (\emph{con} stance) via textual comments. Pro- and con-opinions together form the opinion corpus. 
The data used in our experiments concerns the COVID-19 regulations in the Netherlands during the height of the pandemic, in May 2020. We chose this scenario because 
\begin{enumerate*}[label=(\arabic*)]
    \item we had access to a unique dataset of citizen-provided comments on COVID-19 regulations,
    \item we were able to run the study while the topic was still relevant, making it interesting for crowd workers,
    \item a manual analysis had been performed over the exact same data, allowing for comparison to a human-only baseline, and
    \item the data is reflective of real-world conditions, e.g. feedback was obtained in a matter of days but contains input from a broad group of citizens encompassing broad demographics.
\end{enumerate*}
We analyze feedback from 26,293 Dutch citizens on three of these policy options, treating comments on each option as an opinion corpus. Table~\ref{tab:data-example} shows examples of opinions provided for each different policy option. In our experiments, the \hyena method is applied to one corpus at a time. Since we use data from a publicly run citizen feedback experiment, we observe that some options attracted more pro comments than others. We picked these three options with different pro/con ratios to investigate their impact on the key argument extraction task. The opinions in these corpora are similar to noisy user-generated web comments \shortcite{habernal2017argumentation}, may span multiple sentences, and contain more than one argument at a time. For each policy option, we use the keyword in uppercase as the option identifier in the remainder of the paper.

\begin{table}[tb]
    \centering
    \begin{tabular}{
    @{}
    p{\dimexpr 0.33\linewidth-\tabcolsep}
    p{\dimexpr 0.37\linewidth-2\tabcolsep}
    >{\centering\arraybackslash}p{\dimexpr 0.16\linewidth-2\tabcolsep}
    >{\centering\arraybackslash}p{\dimexpr 0.14\linewidth-\tabcolsep}
    @{}}
        \toprule
        \textbf{Policy option} & \textbf{Example opinion} & \textbf{Num. Opinions} & \textbf{Pro/Con Ratio} \\ 
        \midrule
        \young people may come together in small groups & Then they can go back to school (Pro) & 13400 & 0.66/0.34\\
        All restrictions are lifted for persons who are \immune & Encourages inequality (Con) & 10567& 0.17/0.83\\
        \reopen hospitality and entertainment industry & The economic damage is too high (Pro) & 12814 & 0.55/0.45\\
        \bottomrule
    \end{tabular}
    \caption{Example opinions in the COVID-19 corpora. The collection of opinions for a policy option forms an opinion corpus.}
    \label{tab:data-example}
\end{table}

The original opinions were provided in Dutch. To accommodate a diverse set of annotators in our experiments, we translated all comments to English using the Microsoft Azure Translation service.
All experiments are performed with the translated opinions. Mixing (pretrained) embeddings and machine-translated comments has a minimal impact on downstream task performance \shortcite{sennrich2016improving,eger2018cross,daza2020x}. Although all experiments are conducted in English, the link to the original Dutch text is preserved for future applications.

\subsection{Key Argument Annotation}
\label{sec:hyena-phase-1}
In the first phase of \hyena, human annotators extract individual key argument lists by analyzing the opinion corpus. Since a realistic corpus consists of thousands of opinions, it is unfeasible for an annotator to read all opinions. Thus, \hyena proposes a fixed number of opinions to each annotator. \hyena employs NLP and a sampling technique to select diverse opinions to present to an annotator.

\paragraph{Intelligent Opinion Sampling}
Each annotator is presented, one at a time, with a fixed number of opinions. To sample the next opinion, we embed all opinions and arguments observed thus far using the S-BERT model ($M_S$) \shortcite{reimers2019sentence}. S-BERT converts sentences into fixed-length embeddings, which can be used to compute semantic similarities between pairs of sentences.

Then, we select a pool of candidate opinions using the Farthest-First Traversal (FFT) algorithm \shortcite{Basu2004}. FFT selects the candidate pool as the $f$ farthest opinions in the embedding space from the previously read opinions and annotated arguments (in our experiments, we empirically select $f=5$). Next, we use an argument quality classifier trained on the ArgQ dataset \shortcite{gretz2020large} to select one single clearest opinion related to the policy option. In this way, we aim to increase both the diversity and quality of the opinions presented to each annotator.

\paragraph{Annotation}
Upon reading an opinion, the annotator is asked, first, to \textit{identify} whether the opinion contains an argument or not. If so, the annotator is asked to check whether the argument is already included in their current list of key arguments. If it is not, the annotator should \textit{extract} the argument into a standalone expression (i.e., into a key argument), and add it to the list of key arguments. When adding a new argument, the annotator is asked to indicate the \textit{stance} of the opinion (i.e., whether it is in support or against the related policy option). To facilitate this task, \hyena highlights the most probable stance for the user as a label suggestion \shortcite{schulz2019analysis,beck2021investigating}.

\paragraph{Topic Assignment}
We use a BERTopic \shortcite{grootendorst2020bertopic} model $\mathcal{T}$ to extract clusters of topics from the corpus.  
We train $\mathcal{T}$ on all opinions in the corpus and select the most frequent topics found by $\mathcal{T}$, with duplicates and unintelligible topics manually removed by two experts. We ask human annotators to associate the topics from the generated shortlist with each argument, resulting in an n-hot vector for each argument $a$ per annotator. We obtain the final topic assignment $T$ by summing over all annotators. This topic assignment $T$ is used in the second phase to compute argument similarity. Thus, in the first phase, \hyena yields multiple key argument lists (one per annotator), each containing key arguments and their stances, and an assignment of pre-selected topics to key arguments.

\subsection{Key Argument Consolidation}
In the first phase,
\begin{enumerate*}[label=(\arabic*)] 
    \item the annotators are exposed to a small subset of the opinions in the corpus, and
    \item the interpretation of arguments is subjective.
\end{enumerate*}
In the second phase, \hyena seeks to \emph{consolidate} the key argument lists generated in the first phase.
Our goal is to increase the diversity of the resulting arguments and compensate for individual biases.

First, we create the union of all lists of key arguments generated in the first phase of \hyena.
Then, we ask the annotators to evaluate the similarity of the key argument pairs in the union list. Based on the similarity labels, we employ a clustering algorithm to group similar key arguments, producing a consolidated list of key arguments.

\paragraph{Pairwise Annotation}
To simplify the consolidation task, the annotators are presented with one pair of key arguments at a time and asked whether the concepts described by the key arguments in the pair are semantically similar.
To reduce human effort, we select only the most informative key argument pairs for manual annotation and automatically annotate the remaining pairs. To select the most informative pairs, we adopt a Partial-Ordering approach, \textsc{Power} \shortcite{chai2016cost}, as described below.

Let $p_{ij}$ be a pair of key arguments $\langle a_i, a_j\rangle$. The similarity between the two key arguments in the pair is described by two \textit{similarity scores}, $s^1_{ij}$ and $s^2_{ij}$. By using multiple scores, we seek to make the similarity computation robust.
For each $p_{ij}$, we compute the two similarity scores described in Table~\ref{tab:power-similarity}. We use cosine similarity for $s^1_{ij}$ since the angular distance describes the semantic textual similarity between two arguments. In contrast, we use Euclidean distance for $s^2_{ij}$ since the absolute values of the topic assignment are relevant.

\begin{table}[tb]
    \centering
    
    \begin{tabular}{@{}p{\dimexpr 0.25\linewidth-\tabcolsep}p{\dimexpr 0.75\linewidth-\tabcolsep}@{}}
    \toprule
        \textbf{Measure} & \textbf{Description} \\
    \midrule
        $s^1_{ij} = \frac{\mathbf{i} \cdot \mathbf{j}}{\Vert\mathbf{i}\Vert \Vert\mathbf{j}\Vert}$ & Cosine similarity between embeddings $\mathbf{i} = M_S(a_i)$ and $\mathbf{j} = M_S(a_j)$\\
        $s^2_{ij} = \frac{1}{d(T(a_i),T(a_j))}$ & Inverse of the Euclidean distance $d$ between manual topic assignments $T$ of $a_i$ and $a_j$ \\
    \bottomrule
    \end{tabular}
    \caption{The similarity scores between key argument pairs used to create the pairwise dependency graph.}
    \label{tab:power-similarity}
\end{table}

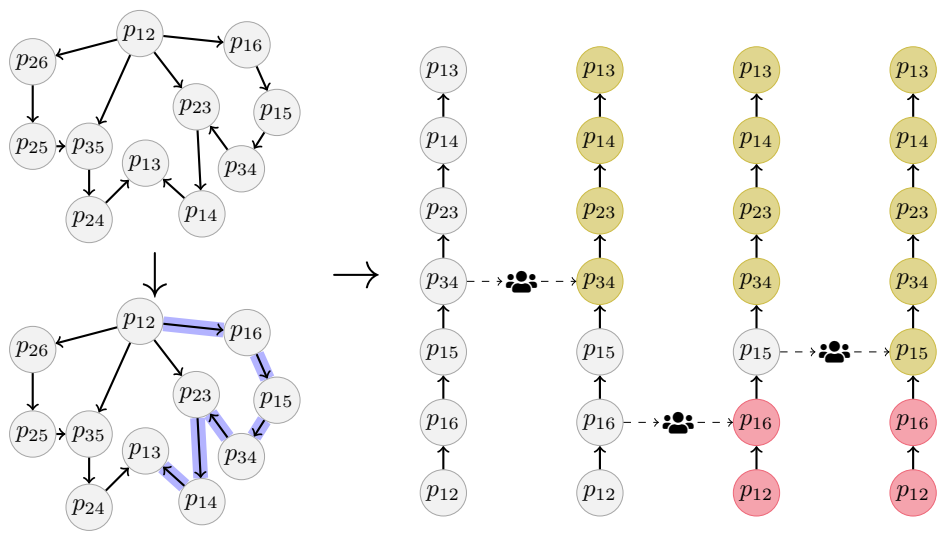
\begin{figure}
\centering
\small
\tikzset{
    table nodes/.style={
        rectangle,
        draw=black,
        align=center,
        minimum height=7mm,
        text depth=0.5ex,
        text height=2ex,
        inner xsep=0pt,
        outer sep=0pt
    },      
    table/.style={
        matrix of nodes,
        row sep=-\pgflinewidth,
        column sep=-\pgflinewidth,
        nodes={
            table nodes
        },
        execute at empty cell={\node[draw=none]{};}
    }
}

\tikzset{path1/.append style={draw=none, draw=blue!70,fill=blue!10}}
\tikzset{path2/.append style={draw=orange!70,fill=orange!10}}
\tikzset{path3/.append style={draw=teal!70,fill=teal!10}}

\tikzset{positive/.append style={draw=paultol4,fill=paultol4!60}}
\tikzset{negative/.append style={draw=paultol2,fill=paultol2!60}}

\tikzset{edge/.style = {->,
        preaction={
            draw,blue!30,-,
            double=blue!30,
            double distance=5\pgflinewidth,
        },
        line width=0.3mm
    }}

\tikzset{thick/.style = {line width=0.3mm}}

\begin{tabular}[t]{ccc}
\begin{tikzpicture}[bn/.style={circle,draw=gray!70,fill=gray!10,minimum size=2mm,inner sep=0.5mm}, every node/.append style={bn},scale=0.8]
    \path   node (1) at  (0, 2)       []  {$p_{12}$}
            node (2) at  (0.1, -0.3)  []  {$p_{13}$}
            node (3) at  (1.1, -1.2)  []  {$p_{14}$}
            node (4) at  (2.43, 0.6)  []  {$p_{15}$}
            node (5) at  (1.9, 1.8)   []  {$p_{16}$}
            node (6) at  (1.0, 0.7)   []  {$p_{23}$}
            node (7) at  (-0.9, -1.3) []  {$p_{24}$}
            node (8) at  (-1.9, 0)    []  {$p_{25}$}
            node (9) at  (-1.9, 1.5)  []  {$p_{26}$}
            node (10) at (1.8, -0.4)  []  {$p_{34}$}
            node (11) at (-0.9, 0)    []  {$p_{35}$};
    \draw[->, line width=0.3mm] (3) edge (2) (1) edge (5) (6) edge (3) (10) edge (6) (4) edge (10) (5) edge (4);
    \draw[->, line width=0.3mm] (7) edge (2) (1) edge (11) (11) edge (7) (8) edge (11);
    \draw[->, line width=0.3mm] (1) edge (9) (9) edge (8);
    \draw[->, line width=0.3mm] (1) edge (6);

\end{tikzpicture} 
& & 
\multirow{3}{*}[2.5cm]{
\begin{tikzpicture}[node distance=0.33cm and 0.5cm, bn/.style={circle,draw=gray!70,fill=gray!10,minimum size=4mm,inner sep=0.5mm},every node/.append style={bn}, scale=0.8]
    \node[] (center)                            {$p_{34}$};
    \node[] (above_1)      [above=of center]    {$p_{23}$};
    \node[] (above_2)      [above=of above_1]   {$p_{14}$};
    \node[] (above_3)      [above=of above_2]   {$p_{13}$};
    \node[] (below_1)      [below=of center]    {$p_{15}$};
    \node[] (below_2)      [below=of below_1]   {$p_{16}$};
    \node[] (below_3)      [below=of below_2]   {$p_{12}$};
    \node[draw=none,fill=none,inner sep=0mm] (ann_1) [right=of center] {\faUsers};
    \node[positive] (center_1)  [right=of ann_1]    {$p_{34}$};
    \node[positive] (above_11)  [above=of center_1] {$p_{23}$};
    \node[positive] (above_21)  [above=of above_11] {$p_{14}$};
    \node[positive] (above_31)  [above=of above_21] {$p_{13}$};
    \node[] (below_11)     [below=of center_1]      {$p_{15}$};
    \node[] (below_21)     [below=of below_11]      {$p_{16}$};
    \node[] (below_31)     [below=of below_21]      {$p_{12}$};
    \node[draw=none,fill=none,inner sep=0mm] (ann_2) [right=of below_21] {\faUsers};
    \node[negative] (below_22)  [right=of ann_2]    {$p_{16}$};
    \node[] (below_12)     [above=of below_22]      {$p_{15}$};
    \node[negative] (below_32)  [below=of below_22] {$p_{12}$};
    \node[positive] (center_2)  [above=of below_12] {$p_{34}$};
    \node[positive] (above_12)  [above=of center_2] {$p_{23}$};
    \node[positive] (above_22)  [above=of above_12] {$p_{14}$};
    \node[positive] (above_32)  [above=of above_22] {$p_{13}$};
    \node[draw=none,fill=none,inner sep=0mm] (ann_3) [right=of below_12] {\faUsers};
    \node[positive] (below_13)  [right=of ann_3] {$p_{15}$};
    \node[negative] (below_23)  [below=of below_13] {$p_{16}$};
    \node[negative] (below_33)  [below=of below_23] {$p_{12}$};
    \node[positive] (center_3)  [above=of below_13] {$p_{34}$};
    \node[positive] (above_13)  [above=of center_3] {$p_{23}$};
    \node[positive] (above_23)  [above=of above_13] {$p_{14}$};
    \node[positive] (above_33)  [above=of above_23] {$p_{13}$};
    \draw [->, thick] (center) -- (above_1);
    \draw [->, thick] (above_1) -- (above_2);
    \draw [->, thick] (above_2) -- (above_3);
    \draw [<-, thick] (center) -- (below_1);
    \draw [<-, thick] (below_1) -- (below_2);
    \draw [<-, thick] (below_2) -- (below_3);
    \draw [->, dashed] (center) -- (ann_1);
    \draw [->, dashed] (ann_1) -- (center_1);
    \draw [->, thick] (center_1) -- (above_11);
    \draw [->, thick] (above_11) -- (above_21);
    \draw [->, thick] (above_21) -- (above_31);
    \draw [<-, thick] (center_1) -- (below_11);
    \draw [<-, thick] (below_11) -- (below_21);
    \draw [<-, thick] (below_21) -- (below_31);
    \draw [->, dashed] (below_21) -- (ann_2);
    \draw [->, dashed] (ann_2) -- (below_22);
    \draw [->, thick] (center_2) -- (above_12);
    \draw [->, thick] (above_12) -- (above_22);
    \draw [->, thick] (above_22) -- (above_32);
    \draw [<-, thick] (center_2) -- (below_12);
    \draw [<-, thick] (below_12) -- (below_22);
    \draw [<-, thick] (below_22) -- (below_32);
    \draw [->, dashed] (below_12) -- (ann_3);
    \draw [->, dashed] (ann_3) -- (below_13);
    \draw [->, thick] (center_3) -- (above_13);
    \draw [->, thick] (above_13) -- (above_23);
    \draw [->, thick] (above_23) -- (above_33);
    \draw [<-, thick] (center_3) -- (below_13);
    \draw [<-, thick] (below_13) -- (below_23);
    \draw [<-, thick] (below_23) -- (below_33);
\end{tikzpicture}
}
\\
{\huge $\downarrow$} & {\huge $\rightarrow$} &\\
\begin{tikzpicture}[node distance=0.6cm, bn/.style={circle,draw=gray!70,fill=gray!10,minimum size=2mm,yshift=3cm,inner sep=0.5mm}, every node/.append style={bn},scale=0.8]
    \path   node (1) at  (0, 2)       []  {$p_{12}$}
            node (2) at  (0.1, -0.3)  []  {$p_{13}$}
            node (3) at  (1.1, -1.2)  []  {$p_{14}$}
            node (4) at  (2.43, 0.6)  []  {$p_{15}$}
            node (5) at  (1.9, 1.8)   []  {$p_{16}$}
            node (6) at  (1.0, 0.7)   []  {$p_{23}$}
            node (7) at  (-0.9, -1.3) []  {$p_{24}$}
            node (8) at  (-1.9, 0)    []  {$p_{25}$}
            node (9) at  (-1.9, 1.5)  []  {$p_{26}$}
            node (10) at (1.8, -0.4)  []  {$p_{34}$}
            node (11) at (-0.9, 0)    []  {$p_{35}$};
    \draw[edge] (3) -- (2);
    \draw[edge] (1) -- (5);
    \draw[edge] (6) -- (3);
    \draw[edge] (10) -- (6);
    \draw[edge] (4) -- (10);
    \draw[edge] (5) -- (4);
    \draw[->, line width=0.3mm] (7) edge (2) (1) edge (11) (11) edge (7) (8) edge (11);
    \draw[->, line width=0.3mm] (1) edge (9) (9) edge (8);
    \draw[->, line width=0.3mm] (1) edge (6);
    
\end{tikzpicture}
& & \\
\end{tabular}
\caption{Pairwise annotation of the dependency graph, combining human and automatic judgments. Vertices indicate argument pairs; the edge direction points to the argument pair with greater similarity. The highlighted blue edges are a disjoint path selected by the \textsc{Power} algorithm. Iteratively, vertices are annotated as similar (yellow) or non-similar (red).}
\label{fig:pairwise}
\end{figure}

Given the similarity scores, we construct a dependency graph $G$ (as in the top-left part of Figure~\ref{fig:pairwise}), where each key argument pair is a vertex in $G$ and the edges indicate a Pareto dependency ($\succ$) between two pairs---the direction of the edge points to the argument pair with greater similarity. A Pareto dependency holds if one of the two scores is strictly greater, with all others being at least equal between two arguments. We define the dependency as follows:
\begin{align}
    \label{eq:pareto0}
    p_{ij} & \succeq p_{i'j'} & \text{if}  & \quad \forall n \quad s^n_{ij} \geq s^n_{i'j'}\\
    \label{eq:pareto1}
    p_{ij} & \succ p_{i'j'}   & \text{if}  & \quad p_{ij} \succeq p_{i'j'} \quad\text{and} \quad \exists n\quad s^n_{ij} > s^n_{i'j'}
\end{align}

Next, we follow \textsc{Power} to extract disjoint paths from $G$. The highlighted path in the bottom-left part of Figure~\ref{fig:pairwise} is an example disjoint path. 
For every path, we perform a pairwise annotation as in the right part of Figure~\ref{fig:pairwise}. We select the vertex at the middle of the unlabeled portion of the path and ask multiple (7) humans to indicate whether the concepts described by the two arguments in the pair are similar on a binary scale, and select the label with the majority vote. Given the annotation, we can automatically label
\begin{enumerate*}[label=(\arabic*)]
    \item all following pairs in the path as similar (yellow) in case the vertex is labeled as similar or
    \item all preceding pairs in the path as non-similar (red) in case the vertex is labeled as non-similar.
\end{enumerate*}
In essence, using the Pareto dependency, we search for threshold similarity scores for each path, above which all pairs are considered similar, and below which all pairs are non-similar. Because this is a local threshold, we prevent over-generalization. To annotate the complete graph efficiently, we employ the parallel Multi-Path annotation algorithm \shortcite{chai2016cost}.

\paragraph{Clustering}
Given a similarity label for each key argument pair, our goal is to identify groups of similar key arguments. However, the similarity among key arguments may not be transitive---given $\langle a_1, a_2\rangle$ as similar and $\langle a_2, a_3\rangle$ as similar, $\langle a_1, a_3\rangle$ may be labeled as dissimilar. This can happen because
\begin{enumerate*}[label=(\arabic*)]
    \item the interpretation of similarity can be subjective (for manually labeled pairs), and
    \item the automatic approach is not always accurate (for automatically labeled pairs).
\end{enumerate*}
Thus, we employ a clustering algorithm for identifying a consolidated list. First, we construct a similarity graph, where each key argument is a vertex and there is an edge between two arguments if they are labeled as similar. Then, we employ out-of-the-box graph clustering algorithms for constructing argument clusters. These clusters form the \emph{key argument lists}.

\subsection{Key Argument Selection}
In the third step of \hyena, we extract a single argument from each cluster, obtaining the final list of key arguments for the opinion corpus. Formally, for every cluster $k \in K$, we create an argument $a_k$ that is \emph{representative} of that cluster. Argument selection methods can be extractive (select an argument from the cluster) or abstractive (generate a new argument that summarizes the cluster). Since there are many methods available for selecting arguments, we can experiment with multiple, and pick the best-performing method. In that case, we again pick an intermediate evaluation metric, which we use to select the best selection method. While there is no human annotation involved in this step, we still consider this higher-level algorithmic design a hybrid process, and thus a collaboration between humans and AI. For the task of selecting relevant arguments, we compare the following four types of approaches.

\begin{description}[topsep=0pt, itemsep=0pt, leftmargin=0pt]
    \item [Centroids] For every cluster $k$, we compute a sentence embedding of every argument $a_k$ using $M_{S}$. Then, we compute pairwise distances between all arguments inside the same cluster. We select the argument with the lowest average distance, measured using cosine similarity, to all other arguments.
    \item [Argument Quality] We use a model that measures argument quality to select the argument from each cluster with the highest quality. We use the same argument classifier as in the Key Argument Annotation phase, trained on the ArgQ dataset \shortcite{gretz2020large}.
    \item [Prompting] We prompt an LLM to synthesize a single argument out of the arguments provided in the argument cluster \shortcite{brown2020language}. We experiment with an open-source and a closed-source model.
    \item [Random] As a baseline, we randomly select an argument from the cluster to represent the entire argument cluster.
\end{description}

\section{Experimental Setup}
\label{sec:experimental}
We involve 378 Prolific (\url{www.prolific.co}) crowd workers as annotators to evaluate \hyena. We required the workers to be fluent in English, have an approval rate above 95\%, and have completed at least 100 submissions. Our experiment was approved by an Ethics Committee and we received informed consent from each subject. We provide supplemental material, containing instructions provided to the annotators, experiment protocol, experiment data, analysis code, and additional details on the experiment \shortcite{vandermeer2024hyenasuppl}.

Table~\ref{tab:experiment-stats} shows an overview of the tasks in the experiment. First, we ask annotators to perform the \hyena method to generate key argument lists for three corpora. Then, we compare the quality of the obtained lists with lists generated for the same corpora via two baselines. All tasks except topic generation were performed by the crowd workers, with most of the task instances annotated by multiple annotators to investigate the agreement between annotators.

\begin{table}[tb]
    \centering
    
    \begin{tabular}{
        @{}
        p{\dimexpr 0.28\linewidth-\tabcolsep}
        >{\centering\arraybackslash}p{\dimexpr 0.17\linewidth-2\tabcolsep}
        p{\dimexpr 0.18\linewidth-2\tabcolsep}
        >{\centering\arraybackslash}p{\dimexpr 0.20\linewidth-2\tabcolsep}
        >{\centering\arraybackslash}p{\dimexpr 0.17\linewidth-\tabcolsep}
        @{}}
        \toprule
            \textbf{Task} & \textbf{Option} & \textbf{Num. Items} & \textbf{Num. Annotators} & \textbf{Num. Annotators per item}\\
        \midrule
            \multirow{3}{*}{Key argument annotation}    & \young        & 255 \texttt{(O)}      & 5 & \multirow{3}{*}{1}\\ 
                                                                        & \immune      & 255 \texttt{(O)}      & 5 & \\ 
                                                                        & \reopen      & 255 \texttt{(O)}      & 5 & \\[.1cm] 
            Topic generation                                            & all          & 45 \texttt{(T)}       & 2$\dagger$ & 2\\[.1cm] 
            \multirow{3}{*}{Topic assignment}           & \young        & 91 \texttt{(A)}       & 10 & \multirow{3}{*}{5}\\ 
                                                                        & \immune      & 66 \texttt{(A)}       & 5  &\\ 
                                                                        & \reopen      & 69 \texttt{(A)}       & 5  &\\[.1cm] 
            \multirow{3}{*}{Key argument consolidation} & \young        & 1538 \texttt{(A+A)}   & 99 & \multirow{3}{*}{3}\\ 
                                                                        & \immune      & 824 \texttt{(A+A)}    & 57 & \\ 
                                                                        & \reopen      & 940 \texttt{(A+A)}    & 87 & \\[.1cm] 
            \midrule
            \multirow{3}{*}{Key argument evaluation}& \young       & 248 \texttt{(O+A)}    & 42 & \multirow{3}{*}{7}\\ 
                                                                      & \immune      & 193 \texttt{(O+A)}    & 29 & \\ 
                                                                      & \reopen      & 221 \texttt{(O+A)}    & 29 & \\ 
        \bottomrule
    \end{tabular}
    \caption{Overview of the tasks in the experiment. Items to be annotated can be opinions \texttt{(O)}, arguments \texttt{(A)}, topics \texttt{(T)}, or combinations. $\dagger$ denotes expert annotators (authors of this paper). }
    \label{tab:experiment-stats}
\end{table}

\subsection{Phase 1: Key Argument Annotation}
\label{sec:experiment-1}
In the first phase of \hyena, each annotator extracts a key arguments list from an opinion corpus. In each corpus, five annotators annotated 51 opinions each, for a total of 255 opinions per corpus. Of the 51 opinions, the first is selected randomly, and the following 50 are selected by FFT. This number of opinions was empirically selected to make the annotation feasible within a maximum of one hour. We instantiate the S-BERT model $M_S$ using the Huggingface Model Hub\footnote{\url{https://huggingface.co/sentence-transformers/all-MiniLM-L6-v2}}. Since our opinion corpus stems from the PVE procedure, we have explicit labels denoting whether a comment was left in favor (\emph{pro}) or opposing (\emph{con}) a proposed policy, which we leverage for the argument stance label suggestion. For obtaining argument quality scores, we use the IBM API~\shortcite{barhaim2021project} to avoid having to retrain a new model. 

\begin{description} [leftmargin=0em,listparindent=1em]
\item[Topics]
We train a BERTopic model on each opinion corpus, generating 59, 56, and 72 topics for the \young, \immune, and \reopen corpora, respectively. Since the number of resulting topics is too high for the manual assignment of arguments to topics, we curate a short list of topics per corpus. We select the 15 most frequent topics in a corpus and ask two experts, the first two authors, to remove duplicates (i.e., topics covering the same semantic aspect) and rate the clarity (i.e., how well the topic describes a relevant aspect of the discussion in the corpus) of each topic. Unique topics with an average clarity score above 2.5 compose the shortlist of topics. Then, we ask crowd annotators to assign topics to each key argument generated in the first phase of \hyena.

\end{description}

\subsection{Phase 2: Key Argument Consolidation}
\label{sec:exp-phase2}
In the second phase of \hyena, we obtain similarity labels $y(a_i,a_j)$ (1 if similar, 0 if not) for all key argument pairs $\langle a_i, a_j\rangle$---some pairs are labeled by the annotators and others are automatically labeled. Given the similarity labels, we construct an argument similarity graph and cluster the graph to identify a consolidated list of key arguments.

\begin{description} [leftmargin=0em,listparindent=1em]
\item[Clustering]
We experiment with two well-known graph clustering algorithms:
\begin{enumerate*}[label=(\arabic*)]
    \item Louvain clustering \shortcite{blondel2008fast} uses network modularity to identify groups of vertices based on a resolution parameter $r$.
    \item Self-tuning spectral clustering \shortcite{zelnik2004self} uses dimensionality reduction in combination with $k$-means to obtain clusters, where $k$ is the desired number of clusters.
\end{enumerate*} We select the parameters of these algorithms to minimize the error metric $E$ shown in Eq.~\ref{eq:cluster-error}.The metric penalizes clusters having dissimilar argument pairs. That is, for a cluster $k \in K$ and $\forall a_i, a_j \in k$, if $y(a_i, a_j) = 1$, the error for that cluster is 0. If a cluster contains only a single element, we manually set the error for that cluster to 1, to discourage creating single-member clusters. We base $E$ on the homogeneity metric \shortcite{rosenberg2007v}, although we do not have access to the ground truth cluster assignments for each argument. Instead, we assume that if all manually labeled arguments are considered similar, they would have been assigned to a single cluster, resulting in a homogenous cluster.

\begin{gather}
\label{eq:cluster-error}
    E = \frac{1}{|K|}\mathlarger{\sum}_{k \in K} \frac{\sum\limits_{a_i,a_j \in k} \mathbbm{1}_{y(a_i,a_j) = 0} }{\binom{|k|}{2}}
\end{gather}
\end{description}

\subsection{Phase 3: Key Argument Selection}
\label{sec:method-phase-3}
In the third phase, we use a mechanism for selecting single arguments per argument cluster. We experiment with multiple methods and different models for selecting arguments. An overview of the methods used is given in Table~\ref{tab:selection-algo}. Below, we explain the setup for each method, and how we select the best-performing method to be used in the final output for \hyena.

\begin{table}[tb]
    \centering
    \begin{tabular}{
    @{}
    >{\arraybackslash}p{\dimexpr 0.20\linewidth-\tabcolsep}
    >{\arraybackslash}p{\dimexpr 0.20\linewidth-2\tabcolsep}
    >{\centering\arraybackslash}p{\dimexpr 0.20\linewidth-2\tabcolsep}
    >{\centering\arraybackslash}p{\dimexpr 0.20\linewidth-2\tabcolsep}
    >{\centering\arraybackslash}p{\dimexpr 0.20\linewidth-2\tabcolsep}
    @{}}
         \toprule
         \textbf{Method} & \textbf{Model} & \textbf{Type} & \textbf{Open} & \textbf{Size} \\
         \midrule
         Random & -- & extractive &  -- & -- \\ 
         Centroid & S-BERT & extractive &  yes & 22M \\
         Prompting & ChatGPT & abstractive & no& 175B\\
         & Llama & abstractive & yes & 7B\\
         Quality & ArgQ & extractive & no & 125M \\
         \bottomrule
    \end{tabular}
    \caption{Argument selection algorithms.}
    \label{tab:selection-algo}
\end{table}

\paragraph{Prompts} We construct different prompts for the two models to extract the desired argument selection output. \emph{ChatGPT} is an instruction-tuned model and can be prompted to answer questions or follow instructions \shortcite{ouyang2022training}. \emph{Llama} lacks instruction-tuning, and thus requires prompts designed for next-token generation \shortcite{touvron2023llama}. 

\begin{promptbox}[label={prompt:chatgpt}]{ChatGPT}
Consider the context of the COVID-19 pandemic and the following arguments:\\
- Argument 1\\
\vdots\\
- Argument $k$\\
\\
Write a key argument that summarizes the above arguments, and make it short and concise.
\end{promptbox}

\begin{promptbox}[label={prompt:llama}]{Llama}
Consider the context of the COVID-19 pandemic and the following arguments:\\
- Argument 1\\
\vdots\\
- Argument $k$\\
\\
A short and concise key argument that summarizes the above arguments is:
\end{promptbox}

\paragraph{Testing Cluster Coherence} First, we investigate the coherence of the clusters generated in Phase 2 according to each argument selection method, with the intent of measuring how each (automated) method aligns with the results of the first two phases of the (hybrid) \hyena process. In cases of low coherence, semantically different arguments may end up together. Vice versa, in highly coherent clusters, only arguments that are the same are actually put together. While the error metric $E$ (Equation~\ref{eq:cluster-error}) gives an error rate, it is mostly a \emph{comparative} method, designed to select the best clustering method. Whether or not the clusters make sense to a human interpreter remains unclear. As such, we devise a so-called \emph{odd-one-out} task, in which we use the Argument Selection methods for selecting arguments from a triple of arguments. In this triple, two arguments stem from the same cluster, and the third from a different cluster. The task for each argument selection method is to select which is the deviating argument. Here, we expect an adequate method to succeed well beyond random performance. Because Argument Quality is not intended for pairwise comparisons of arguments, we omit it in the odd-one-out task. We evaluate the remaining methods on a sample of 1K triples uniformly chosen from all possible triple combinations.

\paragraph{Evaluating Argument Selection} We use different methods and different models for experimenting with the argument selection phase. As before, we employ an error metric to select the best-performing method, which we then inspect through a human evaluation. We use BERT score \shortcite{zhang2020BERTScore}, a metric designed for model selection that uses a trained BERT model to compare the semantic similarity between the selected argument and the original opinions. Specifically, BERT score recall correlates well with human \emph{consistency} judgments, the factual alignment between selected argument and references (original opinions) \shortcite{fabbri2021summeval}. We pick the best-performing method for argument selection based on this metric. This way, we penalize any possible effect of hallucinations of LLMs on the \hyena method. We take the argument selected by each approach in the Key Argument Selection phase of the \hyena procedure. As references, we take all comments that were involved in the creation of the cluster. We compute BERTScore and compare it across our approaches.

\subsection{Baselines}
We compare the output of \hyena to the results of an automated and a manual approach to key argument extraction.

\subsubsection{Comparison to Automated Baseline}
We use the \textbf{\argkp} argument matching model \shortcite{bar2020quantitative} to automatically extract key points from the corpus. \argkp selects candidate key points from opinions using a manually-tuned heuristic, which filters opinions on their length, form, and predicted argument quality \shortcite{gretz2020large}.
The original approach suggests relaxing heuristic parameters such that 20\% of the opinions are selected as candidates. However, this caused overly specific arguments as candidates. Instead, we departed from the parameters used for the ArgKP dataset \shortcite{bar2020quantitative}, and only relax them slightly such that $\sim$10\% of opinions are selected as key point arguments.

Candidate key points and opinions are assigned a match score using a model trained for matching arguments based on RoBERTa \shortcite{liu2019roberta}. Opinions only match the highest-scoring candidate key points if their match score exceeds a threshold $\theta$, corresponding to the best match and threshold (BM+TH) approach. After deduplication, this results in a single list of key arguments per option. We use three metrics, \emph{coverage} ($C$), \emph{precision} ($P$), and \emph{diversity} ($D$) to compare \hyena and \argkp. 

\paragraph{Coverage} ($C$) is defined as the fraction of opinions mapped to an argument out of all the processed opinions \shortcite{bar2020quantitative}. To compute $C$, first, we extract the set of key arguments $\mathcal{A}_H$ from \hyena based on opinions $O_H^{obs}$ ($\subset O$) observed by the annotators. Further, if an argument is extracted from an observed opinion $o_i \in O_H^{obs}$, we add $o_i$ to the set of \emph{annotated} opinions $O^{ann}_H$. Similarly, we extract the set of key arguments $\mathcal{A}_A$ from \argkp based on its observed set of opinions $O_A^{obs} (\equiv O)$, producing a set of \emph{annotated} opinions $O_A^{ann}$. Then, the coverage with respect to \emph{all} observed opinions is:

\begin{equation}
    \label{eq:coverage-hyena-all}
    C_H = \frac{|O^{ann}_H|}{|O^{obs}_{H}|}
\end{equation}

\begin{equation}
    \label{eq:coverage-argkp-all}
    C_A = \frac{|O^{ann}_A|}{|O^{obs}_{A}|}
\end{equation}

Comparing the coverage scores as defined above naively may not be fair since the set of observed opinions (i.e., the denominators of Equations~\ref{eq:coverage-hyena-all} and~\ref{eq:coverage-argkp-all}) are not the same for \hyena and \argkp. Thus, we also compute coverage with respect to a set of \emph{common} opinions, $O^{obs}_{H} \cap O^{obs}_{A}$, observed by both methods, as:

\begin{equation}
\label{eq:coverage-hyena-common}
C_H^{common} = \frac{|O^{ann}_H \cap O^{obs}_{A}|}{|O^{obs}_{H} \cap O^{obs}_{A}|}
\end{equation}

\begin{equation}
\label{eq:coverage-argkp-common}
C_A^{common} = \frac{|O^{ann}_A \cap O^{obs}_{H}|}{|O^{obs}_{H} \cap O^{obs}_{A}|}
\end{equation}

\noindent We add the same term to both denominator and numerator in Equations~\ref{eq:coverage-hyena-common} and~\ref{eq:coverage-argkp-common} so that the coverage stays in the range [0, 1]. Note that $C_H^{common} = C_H$ since $O_H^{obs}, O_H^{ann} \subset O_A^{obs} (\equiv O)$.

\paragraph{Precision} ($P$) is the fraction of mapped opinions for which the mapping is correct \shortcite{bar2020quantitative}. 
Thus, we must map a set of opinions to arguments in order to compute precision. For this mapping, we select the common opinions, $O^{ann}_{H} \cap O^{ann}_{A}$, that are annotated in both \hyena and \argkp. Then for each $o_i \in O^{ann}_{H} \cap O^{ann}_{A}$, we create two pairs $\langle o_i, \mathcal{A}_H(o_i)\rangle$ and $\langle o_i, \mathcal{A}_A(o_i)\rangle$, where $\mathcal{A}_H(o_i)$ and $\mathcal{A}_A(o_i)$ are the arguments associated with $o_i$ by \hyena and \argkp, respectively. Then, we ask annotators to label $z(o_i, a_i)=1$ for all matching pairs and $z(o_i, a_i)=0$ for all non-matching pairs, and keep the majority consensus from multiple annotators. Given the opinion-argument mapping, we compute precision as:

\begin{equation}
\centering
    P_H^{common} = \frac{\sum\limits_{o_i \in O_H^{ann} \cap O_A^{ann}}z(o_i, \mathcal{A}_H(o_i))}{|O_H^{ann} \cap O_A^{ann}|} \label{eq:precision-same-hyena}
\end{equation}

\begin{equation}
\centering
    P_A^{common} = \frac{\sum\limits_{o_i \in O_H^{ann} \cap O_A^{ann}}z(o_i, \mathcal{A}_A(o_i))}{|O_H^{ann} \cap O_A^{ann}|} \label{eq:precision-same-argkp}
\end{equation}

\paragraph{Diversity} ($D$) is defined as the ratio of key arguments and the number of comments seen by the method. We use diversity to signify how well our method is able to preserve the perspectives present in the opinions seen by the method. In order to compare across methods, we take (1) only correct mappings ($z(o_i, a_i)=1$) using the labels from $P$ and (2) take the opinions seen by both $A$ and $H$. We define diversity as follows:

\begin{equation}
\centering
    D_H = \frac{\mathcal{A}_H}{|O^{obs}_{H} \cap O^{obs}_{A}|}
\end{equation}
\begin{equation}
\centering
    D_A = \frac{\mathcal{A}_A}{|O^{obs}_{H} \cap O^{obs}_{A}|}
\end{equation}

\subsubsection{Comparison to Manual Baseline}
A manual analysis involving six experts examined a portion of the feedback stemming from the PVE procedure. This analysis included a sample of participants (2,237 out of 26,293) for identifying key arguments \shortcite{mouter2021public}, where each expert generated a list of arguments for and against each of the relaxation measures based on the opinion text. A single participant could leave multiple opinions, and the analysis does not report the exact number of opinions analyzed. Since we have access to 36,781 opinions for the three options (Table~\ref{tab:data-example}), we estimate the number of opinions the six experts would have analyzed to be 3,129 across the three options (following each participant entering $\pm 1.4$ opinions), and at least 2,237 (at least one opinion per participant). In contrast, \hyena annotators analyze 765 intelligently selected opinions across the three options. 

It is evident that \hyena reduces the number of opinions analyzed. Further, we investigate the extent to which the key argument lists generated by \hyena and the manual baseline have comparable insights. To do so, we report the number of \hyena key arguments that are overlapping, missing, and new compared to the expert-identified key arguments. We cannot compute precision and coverage for the manual baseline because it does not include a mapping between key arguments and opinions.

\section{Results}
\label{sec:results}

First, we analyze the inter-rater reliability of annotations. Then, we analyze the intermediate results of the three phases of \hyena. Finally, we compare our hybrid approach with the automated and manual baselines. 

\subsection{Annotator Agreement}
Table~\ref{tab:results-raters} shows the inter-rater reliability (IRR) for four steps with overlapping human annotations. We didn't obtain IRR ratings for the argument extraction task in Phase 1 since the annotation is designed to be disjoint, and raters had little to no overlap in their extractions. In the Topic Generation phase (Section 4.1), we use the intraclass correlation coefficient ICC$(3,k)$ \shortcite{shrout1979intraclass} since it involves ordinal ratings. In the other three tasks, multiple binary labels are obtained for the same subjects. In these tasks, we use prevalence- and bias-adjusted $\kappa$ (PABAK) \shortcite{sim2005kappa}, which adjusts Fleiss' $\kappa$ for prevalence and bias resulting from small or skewed distribution of ratings.

\begin{table}[tb]
    \centering
    \begin{tabular}{
    @{}
        p{\dimexpr 0.60\linewidth-\tabcolsep}
        >{\centering\arraybackslash}p{\dimexpr 0.20\linewidth-2\tabcolsep}
        >{\centering\arraybackslash}p{\dimexpr 0.20\linewidth-\tabcolsep}
    @{}}
         \toprule
         \textbf{Task} & \textbf{ICC3k} & \textbf{PABAK} \\
         \midrule
         Topic Generation       & 0.66 (0.14) & -- \\
         Topic Assignment       & -- & 0.81 (0.10)\\
         Key Argument Consolidation & -- & 0.34 (0.03)\\
         Key Argument Evaluation  & -- & 0.36 (0.04)\\
        \bottomrule
    \end{tabular}
    \caption{IRR scores per task in \hyena. We show the average (and standard deviation) over the three option corpora.}
    \label{tab:results-raters}
\end{table}

In Topic Generation, the main source of the disagreement stems from a single option: \textsc{reopen}. Here, the annotators rated two topics almost inverted (rating 4 versus rating 2) out of a 1--5 Likert scale, resulting in an ICC score of 0.46. The two topics contained the words \emph{``mental health income decrease,''} and \emph{``measures rules these should''}. For the other two options, \textsc{young} and \textsc{immune}, a higher score of 0.71 and 0.80 were obtained respectively. 

We obtained the lowest reliability scores for the last two annotation tasks, Key Argument Consolidation and Key Argument Evaluation. The obtained scores may be due to the difficulty of the task---for instance, lay annotators are asked to characterize the similarity between two arguments, and they may not stick to the provided definition of argument similarity. However, task difficulty may not be the only factor at play here. Argument comparisons are made with limited context, and the personal perspective or background of the annotator may influence their judgment. Thus, the low IRR scores may indicate a combination of task difficulty and the relatively subjective nature of the task \shortcite{aroyo2015truth}. Similar reasoning holds for the task of evaluating the match between the extracted argument and the original opinions.

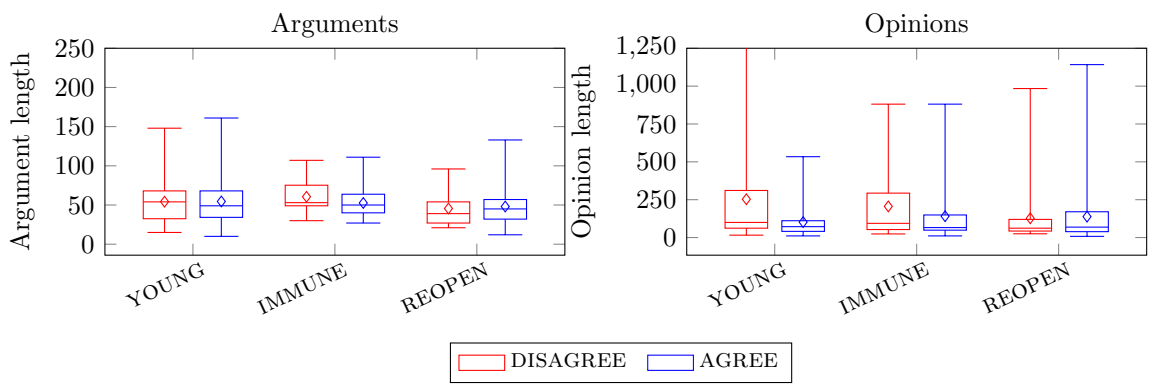
\begin{figure}[tb]
    \centering
     \begin{tikzpicture}
\tikzstyle{every node}=[font=\small]
    \begin{axis}[
    name=plot-motivations,
    boxplot/draw direction=y,
    title={Arguments},
    title style={yshift=-0.2cm},
	ytick distance=50,
	ymax = 250,
    xtick={1,2,3},
	xticklabels={\textsc{young}, \textsc{immune}, \textsc{reopen}},
    xticklabel style={rotate=25, anchor=east, yshift=-0.55em, xshift=0.5em},
	width=0.5\columnwidth,
	height=4.3cm,
	ylabel={Argument length},
    /pgfplots/boxplot/box extend=0.3,
    legend pos = north west,
    legend style={at={(0.75,-0.52)}, anchor=west, /tikz/every even column/.append style={column sep=0.2cm}},
    legend columns=2,
    legend image code/.code={%
      \fill[#1,fill=none] (0cm,-0.1cm) rectangle (0.6cm,0.1cm);
    },
	]
    \addplot+ [
    boxplot prepared={
      upper quartile=68.0,
      lower quartile=32.5,
      upper whisker=148,
      lower whisker=15,
      median=54.0,
      average=54.23255813953488,
      draw position=0.8,
      every average/.style={/tikz/mark=diamond},
    },
    draw=red,
    style=solid,
    ] coordinates {};
    \addplot+ [
        forget plot,
    boxplot prepared={
      upper quartile=75.25,
      lower quartile=49.0,
      upper whisker=107,
      lower whisker=30,
      median=53.0,
      average=60.5,
      draw position=1.8,
      every average/.style={/tikz/mark=diamond},
    },
    draw=red,
    style={solid},
    ] coordinates {};
    \addplot+ [
    forget plot,
    boxplot prepared={
      upper quartile=54.0,
      lower quartile=27.0,
      upper whisker=96,
      lower whisker=21,
      median=39.0,
      average=45.56,
      draw position=2.8,
      every average/.style={/tikz/mark=diamond},
    },
    draw=red,
    style={solid},
    ] coordinates {};
    
    \addplot+ [
    boxplot prepared={
      upper quartile=68.0,
      lower quartile=34.25,
      upper whisker=161,
      lower whisker=10,
      median=49.0,
      average=54.56349206349206,
      draw position=1.2,
      every average/.style={/tikz/mark=diamond},
    },
    draw=blue,
    ] coordinates {};
    \addplot+ [
    boxplot prepared={
      upper quartile=63.75,
      lower quartile=40.0,
      upper whisker=111,
      lower whisker=27,
      median=50.0,
      average=52.611111111111114,
      draw position=2.2,
      every average/.style={/tikz/mark=diamond},
    },
    draw=blue,
    style={solid},
    ] coordinates {};
    \addplot+ [
    boxplot prepared={
      upper quartile=57.0,
      lower quartile=32.0,
      upper whisker=133,
      lower whisker=12,
      median=45.0,
      average=48.06666666666667,
      draw position=3.2,
      every average/.style={/tikz/mark=diamond},
    },
    draw=blue,
    style={solid},
    ] coordinates {};
    \legend {\textsc{disagree}, \textsc{agree}};
    \end{axis}
    
        \begin{axis}[
    name=plot-arguments,
    anchor=west,
    at={($(plot-motivations.east)+(1.6cm,0cm)$)},
    boxplot/draw direction=y,
    title={Opinions},
    title style={yshift=-0.2cm},
	ytick distance=250,
    xtick={1,2,3},
	xticklabels={\textsc{young}, \textsc{immune}, \textsc{reopen}},
    xticklabel style={rotate=25, anchor=east, yshift=-0.55em, xshift=0.5em},
    width=0.5\columnwidth,
	height=4.3cm,
	ymax=1250,
	ylabel={Opinion length},
    /pgfplots/boxplot/box extend=0.3,
	]
    \addplot+ [
    boxplot prepared={
      upper quartile=311.0,
      lower quartile=62.0,
      upper whisker=1767,
      lower whisker=16,
      median=100.0,
      average=253.13953488372093,
      draw position=0.8,
      every average/.style={/tikz/mark=diamond},
    },
    draw=red,
    style={solid},
    ] coordinates {};
    \addplot+ [
    boxplot prepared={
      upper quartile=293.5,
      lower quartile=53.0,
      upper whisker=881,
      lower whisker=24,
      median=94.0,
      average=205.8181818181818,
      draw position=1.8,
      every average/.style={/tikz/mark=diamond},
    },
    draw=red,
    style={solid},
    ] coordinates {};
    \addplot+ [
    boxplot prepared={
      upper quartile=120.0,
      lower quartile=43.0,
      upper whisker=984,
      lower whisker=25,
      median=62.0,
      average=127.24,
      draw position=2.8,
      every average/.style={/tikz/mark=diamond},
    },
    draw=red,
    style={solid},
    ] coordinates {};
    
    \addplot+ [
    boxplot prepared={
      upper quartile=111.25,
      lower quartile=41.0,
      upper whisker=534,
      lower whisker=11,
      median=72.0,
      average=102.3015873015873,
      draw position=1.2,
      every average/.style={/tikz/mark=diamond},
    },
    draw=blue,
    style={solid},
   ] coordinates {};
    \addplot+ [
    boxplot prepared={
      upper quartile=149.25,
      lower quartile=49.0,
      upper whisker=881,
      lower whisker=11,
      median=66.0,
      average=141.22222222222223,
      draw position=2.2,
      every average/.style={/tikz/mark=diamond},
    },
    draw=blue,
    style={solid},
    ] coordinates {};
    \addplot+ [
    boxplot prepared={
      upper quartile=170.5,
      lower quartile=39.0,
      upper whisker=1142,
      lower whisker=8,
      median=69.0,
      average=138.33333333333334,
      draw position=3.2,
      every average/.style={/tikz/mark=diamond},
    },
    draw=blue,
    style={solid},
    ] coordinates {};
    \end{axis}
    \end{tikzpicture}
    \caption{Disagreement analysis for the Key Argument Evaluation phase. On the left, argument lengths are the same whether annotators agree or disagree. However, on the right, annotators disagree on match labels in long opinions.}
    \label{fig:disagreement-phase3}
\end{figure}

Focusing on the evaluation phase, we compare argument--opinion pairs where large disagreement was observed (\textsc{disagree}) to pairs with low disagreement (\textsc{agree}) in Figure~\ref{fig:disagreement-phase3}. 
Specifically, we compared the lengths of the arguments and opinions. We find that the lengths of the arguments--opinion pairs with large inter-rater disagreement did not differ from those with low disagreement. However, we found considerably longer opinions on average when annotators disagreed. Possibly, long opinions contain multiple arguments, which in turn may cause the annotator to fail to identify the provided argument.

Prolific annotators were generally young (M=29.2, SD=7.8) and typically active users with a median of over 300 tasks completed (M=404, SD=418). A little over half of our annotators were male (58.8\%), another 38.6\% reported as female, and the rest had no data available. 76.7\% reported a language other than English as their native language (we did require all annotators to be fluent in English). Annotators mostly resided in European countries, with the UK, Mexico, and the US being the only non-EU countries with more than 10 annotators. 23.8\% reported as being a full-time student, with the rest either reporting as not being a student or having no data available. Further work is required in order to investigate the impact of demographic factors on the subjective interpretation of the opinions and arguments involved \shortcite{shortall2022reason}.

\subsection{Phase 1: Key Argument Annotation}
In Phase 1, individual annotators were guided through 51 opinions each and asked to annotate the observed arguments.
Table~\ref{tab:experiment1-results-generation} shows the number of different operations annotators perform over the 51 opinions. On average, the annotators identified 15 unique key arguments per option. About half of the opinions were skipped, mainly because the opinion lacked a clear argument. 
Since the opinions had been automatically translated, we also provided annotators with the option to skip an opinion due to an unclear translation. Out of 51 actions, annotators reported mistranslations in 6, 7, and 2 opinions on average for \young, \immune, and \reopen, respectively.

\begin{table}[bt]
    \centering
    \begin{tabular}{
        @{}
        >{\centering\arraybackslash}p{\dimexpr 0.17\linewidth-\tabcolsep}
        >{\centering\arraybackslash}p{\dimexpr 0.17\linewidth-2\tabcolsep}
        >{\centering\arraybackslash}p{\dimexpr 0.17\linewidth-2\tabcolsep}
        >{\centering\arraybackslash}p{\dimexpr 0.17\linewidth-2\tabcolsep}
        >{\centering\arraybackslash}p{\dimexpr 0.17\linewidth-2\tabcolsep}
        >{\centering\arraybackslash}p{\dimexpr 0.17\linewidth-\tabcolsep}
        @{}}
         \toprule
         & \multicolumn{3}{c}{\textbf{Phase 1}} & \multicolumn{2}{c}{\textbf{Phase 2}} \\
         \cmidrule(lr){2-4}\cmidrule(lr){5-6}
         \textbf{Option} & \textbf{\# Args} &\textbf{\# Skip} & \textbf{\# Already} & $\Delta$ & \textbf{$\tau$} \\
         \midrule
         \young   & 18.0 (5.5) & 23.4 (5.4)  & 11.4 (9.0) & -61.6\% & 0.34\\ 
         \immune  & 12.8 (2.6) & 31.4 (4.5)  & 8.6 (4.4)  & -59.1\% & 0.42\\ 
         \reopen & 13.8 (7.6) & 29.2 (11.5) & 10.2 (7.6)  & -59.8\% & 0.41\\ 
         \bottomrule
    \end{tabular}
    \caption{The average annotation operations (and their standard deviation) in Phase 1, and obtained statistics for Phase 2.}
    \label{tab:experiment1-results-generation}
    
\end{table}

This is a positive result since the noise (i.e., irrelevant or non-argumentative opinions) in public feedback can be much higher. Thus, the argument quality classifier we incorporate for opinion sampling is effective in filtering noise. Further, the annotators marked only about 15\% of the encountered opinions as already annotated key arguments, which shows that the FFT approach is effective in sampling a diverse set of opinions for annotation.

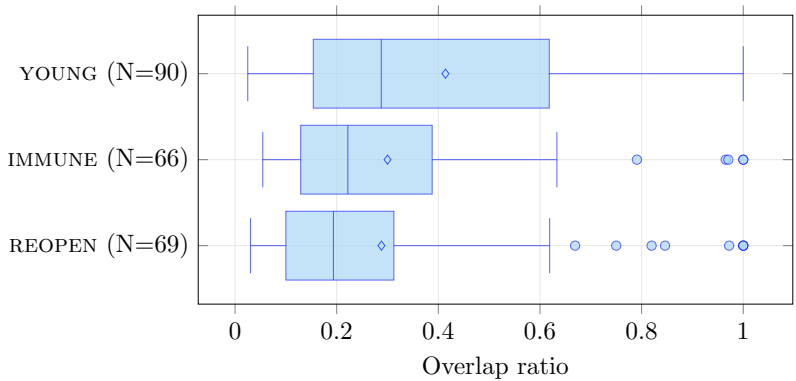
\begin{figure}[bt]
    \centering
    \pgfplotsset{compat=1.16}
\pgfplotstableread{
0.07575757575757576
0.05454545454545454
0.125
0.13043478260869565
0.3076923076923077
0.09090909090909091
0.24390243902439024
0.34782608695652173
0.09210526315789473
0.32558139534883723
0.07142857142857142
0.27102803738317754
0.2786885245901639
0.3157894736842105
0.1643835616438356
0.3225806451612903
0.38461538461538464
0.09836065573770492
0.32
0.0967741935483871
0.08108108108108109
1.0
0.3888888888888889
0.48484848484848486
0.14285714285714285
0.12
0.13043478260869565
0.08163265306122448
0.05660377358490566
0.06306306306306306
0.14545454545454545
0.12903225806451613
0.09090909090909091
0.3333333333333333
0.07142857142857142
0.5
0.9705882352941176
1.0
0.7906976744186046
0.48148148148148145
0.30357142857142855
0.30158730158730157
0.3088235294117647
0.1276595744680851
0.46511627906976744
0.15384615384615385
0.5384615384615384
0.5384615384615384
0.25
0.6
0.2926829268292683
0.12962962962962962
0.14705882352941177
0.9655172413793104
0.4745762711864407
0.19047619047619047
0.16326530612244897
0.2
0.1956521739130435
0.5490196078431373
0.6333333333333333
0.140625
0.19117647058823528
0.19047619047619047
0.136986301369863
0.4594594594594595
}\immunetable

\pgfplotstableread{
0.4186046511627907
0.56
0.14285714285714285
0.14705882352941177
0.42857142857142855
0.10810810810810811
0.5333333333333333
1.0
0.06451612903225806
0.17857142857142858
0.16
0.14285714285714285
0.6
0.1875
0.35294117647058826
0.3409090909090909
0.5
0.7142857142857143
0.06
0.2682926829268293
0.1282051282051282
0.2553191489361702
0.24324324324324326
0.0967741935483871
0.16831683168316833
0.6666666666666666
0.625
0.2
0.3125
0.6086956521739131
0.42857142857142855
0.14782608695652175
0.14583333333333334
0.2876712328767123
0.20689655172413793
0.1388888888888889
0.5897435897435898
0.21428571428571427
0.5238095238095238
0.18012422360248448
0.26
0.21739130434782608
0.21052631578947367
0.12903225806451613
0.25
0.1702127659574468
0.6
0.7307692307692307
0.3473684210526316
0.17567567567567569
0.09523809523809523
0.04081632653061224
0.24561403508771928
0.23728813559322035
0.19642857142857142
0.3170731707317073
0.2987012987012987
0.07751937984496124
0.6222222222222222
0.3888888888888889
0.9782608695652174
0.46987951807228917
1.0
0.6585365853658537
0.48214285714285715
0.5094339622641509
1.0
0.9344262295081968
0.9583333333333334
0.9761904761904762
0.7
0.875
0.9411764705882353
1.0
1.0
0.7162162162162162
1.0
1.0
1.0
0.6142857142857143
0.04411764705882353
0.13513513513513514
0.21739130434782608
0.025
0.05172413793103448
0.10256410256410256
0.12
1.0
0.1111111111111111
0.21052631578947367
0.11764705882352941
}\youngtable

\pgfplotstableread{
0.05128205128205128
0.09090909090909091
0.08333333333333333
0.08571428571428572
0.1
0.225
0.06666666666666667
0.08823529411764706
0.15384615384615385
0.07692307692307693
0.04878048780487805
0.10714285714285714
0.04878048780487805
0.2916666666666667
0.09523809523809523
0.2727272727272727
0.2777777777777778
0.11538461538461539
0.125
0.3125
0.18518518518518517
0.25
0.2857142857142857
0.22916666666666666
0.2978723404255319
0.14516129032258066
0.08571428571428572
0.5652173913043478
0.06666666666666667
0.75
0.16666666666666666
0.24444444444444444
0.13513513513513514
0.046511627906976744
0.21153846153846154
0.05714285714285714
0.2
0.17777777777777778
0.35135135135135137
0.16279069767441862
0.23333333333333334
0.05
0.1935483870967742
0.3
0.078125
0.030303030303030304
0.16216216216216217
0.16071428571428573
0.15625
0.1111111111111111
0.6190476190476191
0.9722222222222222
0.819672131147541
0.30434782608695654
1.0
0.49056603773584906
0.6024096385542169
0.5217391304347826
1.0
0.8461538461538461
0.5
1.0
0.6691729323308271
0.18518518518518517
0.14814814814814814
0.5217391304347826
0.45454545454545453
0.20689655172413793
0.2073170731707317
}\reopentable

\begin{tikzpicture}
    \begin{axis} [ybar,
        name=plot-arg-overlap,
        boxplot/draw direction=x,
        grid=both,
        width=0.7\textwidth,
        height=6cm,
        xlabel={Overlap ratio},
        ytick={1,2,3},
        yticklabels={\reopen (N=69), \immune (N=66),\young (N=90)},
        grid style={line width=.1pt, draw=gray!20},
    ]
    \addplot+[boxplot, draw=bblue!20!blue,fill=bblue!50, opacity=0.7] table[y index=0, row sep=newline]{\reopentable};
    \addplot+[boxplot, draw=bblue!20!blue,fill=bblue!50, opacity=0.7] table[y index=0, row sep=newline]{\immunetable};
    \addplot+[boxplot, draw=bblue!20!blue,fill=bblue!50, opacity=0.7] table[y index=0, row sep=newline]{\youngtable};
    \addplot+[only marks,draw=bblue!50!blue,fill=bblue, mark=diamond] 
       table [
         y=y,
         x=x
        ]
    {
        x y
        0.288 1
        0.300 2
        0.414 3
    };
    \end{axis}
\end{tikzpicture}
    \caption{Distribution of argument overlap ratio for arguments generated by Key Argument Annotation in Phase 1.}
    \label{fig:overlap}
\end{figure}

Our instructions did not include an explicit mention of whether copying from the opinion text was allowed, but we observed that annotators often paraphrased arguments from opinions. To examine the behavior of the annotators, we measured the amount of text that was literally copied from the opinions. To do so, we take the largest common substring on the character level between opinion text and argument and divide it by the length of the argument. In Figure~\ref{fig:overlap}, we show the distribution of overlap ratios across all extracted arguments. While some arguments do get copied verbatim (overlap ratio of 1), across all three corpora annotators generally rephrase the arguments. This shows that, in \hyena, human intervention acts in shaping the arguments extracted from the opinions, rather than simply copying part of an opinion (as automated methods would do). Table~\ref{tab:arg-copy} shows some examples of arguments extracted with different overlap ratios.

\begin{table}[tb]
    \centering
    \begin{tabular}{
    @{}
    >{\centering\arraybackslash}p{\dimexpr 0.10\linewidth-\tabcolsep}
    >{\arraybackslash}p{\dimexpr 0.42\linewidth-2\tabcolsep}
    >{\arraybackslash}p{\dimexpr 0.36\linewidth-2\tabcolsep}
    >{\centering\arraybackslash}p{\dimexpr 0.12\linewidth-\tabcolsep}
    @{}}
    \toprule
         \textbf{Option} & \multicolumn{1}{c}{\textbf{Opinion Text}} & \multicolumn{1}{c}{\textbf{Extracted Argument}} & \makecell[c]{\textbf{Overlap}\\\textbf{Ratio}}\\
    \midrule
        \young & Our daughter misses {\color{green}he}r friends so much and I notice that she really needs it & Positive for the psychological {\color{green}he}alth of children & 0.060\\
        \immune & Keep one system, keep it simple. Not too many devia{\color{green}tions}. & Everyone should be subject to the same set of rules/restric{\color{green}tions.}  & 0.091\\
        \immune & Too little research has been done to limit the measures for people who are immune and too few opportunities to test it. In addition, {\color{green}it is difficult to control.} & {\color{green}It is difficult to control.} & 1.000\\
        \reopen & These measures are quite {\color{green}easy to take compared to the unselected measures.} & Measures are {\color{green}easy to take compared to the unselected measures} & 0.820\\
    \bottomrule
    \end{tabular}
    \caption{Examples of extracted arguments in Phase 1 of \hyena. Overlapping character sequences are highlighted in green.}
    \label{tab:arg-copy}
\end{table}

The topic models for each option generated a large variety of topics. After the generation of the topic models $\mathcal{T}$, we retain only the top-15 most frequent topics to make the annotation feasible. Our experts eliminated one, two, and zero topics as duplicates in the three options (Table~\ref{tab:expert-topic-stats}). On average, the coherence scores---ranging from 1 (low) to 5 (high)---are high. This suggests that these topics were suitable for assignment to the arguments stemming from the crowd-extracted arguments. Table~\ref{tab:expert-topic-overview} shows examples from the final list of topics, with low-scoring topics removed.

\begin{table}[tb]
    \centering
    \begin{tabular}{
    @{}
    >{\centering\arraybackslash}p{\dimexpr 0.20\linewidth-\tabcolsep}
    >{\centering\arraybackslash}p{\dimexpr 0.20\linewidth-2\tabcolsep}
    >{\centering\arraybackslash}p{\dimexpr 0.20\linewidth-2\tabcolsep}
    >{\centering\arraybackslash}p{\dimexpr 0.20\linewidth-2\tabcolsep}
    >{\centering\arraybackslash}p{\dimexpr 0.20\linewidth-2\tabcolsep}
    @{}}
    \toprule
         \textbf{Option} & $|\mathcal{T}|$  & \makecell[b]{\textbf{Number of}\\ \textbf{duplicates}} & \textbf{Kept} & \makecell[b]{\textbf{Average}\\\textbf{rating}}\\
    \midrule
         \young  & 59 & 1 & 12 & 4.4\\
         \immune & 56 & 2 & 12 & 4.4\\
         \reopen & 72 & 0 & 14 & 4.0\\
    \bottomrule
    \end{tabular}
    \caption{Expert topic generation statistics in Phase 1.}
    \label{tab:expert-topic-stats}
\end{table}

\newtcbox{\varbox}{
    colframe=blue!15!white,
    colback=blue!15!white,
    varwidth upper=3.3cm,
    size=fbox,
    before=,
    after=,
    rounded corners=all,
}

\begin{table}[tb]
    \centering
    \small
    \begin{tabular}{
    @{}
    >{\arraybackslash}p{\dimexpr 0.2\linewidth-\tabcolsep}
    >{\centering\arraybackslash}p{\dimexpr 0.2\linewidth-2\tabcolsep}
    >{\centering\arraybackslash}p{\dimexpr 0.6\linewidth-\tabcolsep}
    @{}}
    
    \toprule
         \textbf{Option} & \textbf{Clarity Rating} & \textbf{Topic words} \\
\midrule
\young & 4.5 & immune entertainment hospitality restrictions \\
& 4 & infection immunity risk infected \\
& 4 & \emph{virus susceptible spread transmit} \\
& 4.5 & \emph{schools reopen education students} \\
& 5 & \emph{risk limited low dangerous} \\
& 5 & \emph{group risk target least} \\[.3em]
\immune & 4.5 & homes nursing care vulnerable \\
& 4 & netherlands country provinces dutch \\
& 5 & \emph{risk contamination danger dangerous} \\
& 4.5 & \emph{work companies home economy} \\
& 5 & entertainment hospitality catering industry \\[.3em]
\reopen & 5 & homes nursing care vulnerable \\
& 4 & netherlands friesland groningen dutch \\
& 5 & risk hospitality entertainment dangerous \\
& 3 & \emph{mental health income decrease} \\
& 3 & \emph{measures rules these should} \\
    \bottomrule
    \end{tabular}
    \caption{Examples of topics generated in Phase 1, including the top 4 words and the average clarity rating. Option-specific topics are \emph{emphasized}.}
    \label{tab:expert-topic-overview}
\end{table}

\subsection{Phase 2: Key Argument Consolidation}
In Phase 2, \hyena uses the \textsc{Power} algorithm to guide human annotations on arguments similarity, with the intent of creating clusters of similar arguments across all arguments individually annotated in Phase 1.
Table~\ref{tab:experiment1-results-generation} (right side) shows the benefit of the \textsc{Power} algorithm---the number of pairs requiring human annotation ($\Delta$) was on average reduced by $60\%$. 
The transitivity scores $\tau$ \shortcite{newman2002random} measure the extent to which transitivity holds among the similarity labels of argument pairs. The low $\tau$ scores indicate the need for subsequent clustering, given that there are no clear graph components in which all arguments are similar.  

\begin{figure}[tb]
    \centering
    \begin{tikzpicture}
	\begin{axis}[
	    name=reopen_l,
	    clip=false,
	    title=\textsc{reopen} Louvain,
	    title style={yshift=-0.2cm},
        width=0.5\columnwidth,
        height=3.7cm,
        grid=both,
        grid style={line width=.1pt, draw=gray!20},
        legend columns=3,
        legend style={
            at={(1.13, -0.35)},
            anchor=north,
            /tikz/every even column/.append style={column sep=0.3cm}
        },
        ymax=1,
        xmin=0,
		xlabel=$r$,
		ylabel=$E$,
		ytick distance=0.2,
		xlabel shift=-0.1cm,
		extra y ticks={0.12456140350877193},
        extra tick style={grid=major,major grid style={red,thick},
        tick label style={anchor=east}},
        extra y tick labels={},
        xlabel shift=-0.15cm]
	\addplot[color=blue] file{images/image_data/louvain_cluster_errors_option8.dat};

	\end{axis}
	
	\begin{axis}[
	    name=reopen_r,
	    at={($(reopen_l.east)+(1cm,0)$)},
	    anchor=west,
        title=\textsc{reopen} \textbf{spectral},
        title style={yshift=-0.2cm},
        width=0.5\columnwidth,
        ymax=1,
        height=3.7cm,
		xlabel=$k$,
		ytick distance=0.2,
        grid=both,
        grid style={line width=.1pt, draw=gray!20},
		extra y ticks={0.12456140350877193},
        extra tick style={grid=major,major grid style={red,thick},
        tick label style={anchor=east}},
        extra y tick labels={},
    xlabel shift=-0.15cm]
	\addplot[color=blue] file{images/image_data/spectral_cluster_errors_option8.dat};

	\end{axis}
	
	\begin{axis}[
	    name=immune_l,
	    at={($(reopen_l.west)+(0,3.5cm)$)},
	    anchor=west,
        title=\textsc{immune} \textbf{Louvain},
        title style={yshift=-0.2cm},
        width=0.5\columnwidth,
        ymax=1,
        height=3.7cm,
		xlabel=$r$,
		ylabel=$E$,
        grid=both,
        grid style={line width=.1pt, draw=gray!20},
		ytick distance=0.2,
		xlabel shift=-0.15cm,
		extra y ticks={0.04224664224664224},
        extra tick style={grid=major,major grid style={red,thick},
        tick label style={anchor=east}},
        extra y tick labels={}]
	\addplot[color=blue] file{images/image_data/louvain_cluster_errors_option5.dat};
	\end{axis}
	
	\begin{axis}[
	    name=immune_r,
	    at={($(reopen_l.east)+(1cm,3.5cm)$)},
	    anchor=west,
        title=\textsc{immune }spectral,
        title style={yshift=-0.2cm},
        width=0.5\columnwidth,
        ymax=1,
        height=3.7cm,
		xlabel=$k$,
        grid=both,
        grid style={line width=.1pt, draw=gray!20},
		ytick distance=0.2,
		xlabel shift=-0.15cm,
		extra y ticks={0.04224664224664224},
        extra tick style={grid=major,major grid style={red,thick},
        tick label style={anchor=east}},
        extra y tick labels={}]
	\addplot[color=blue] file{images/image_data/spectral_cluster_errors_option5.dat};
	\end{axis}
	
	\begin{axis}[
	    name=young_l,
	    at={($(immune_l.west)+(0,3.5cm)$)},
	    anchor=west,
        title=\textsc{young} \textbf{Louvain},
        title style={yshift=-0.2cm},
        width=0.5\columnwidth,
        ymax=1,
        height=3.7cm,
        grid=both,
        grid style={line width=.1pt, draw=gray!20},
		xlabel=$r$,
		ylabel=$E$,
		ytick distance=0.2,
		xlabel shift=-0.15cm,
		extra y ticks={0.11059757236227824},
        extra tick style={grid=major,major grid style={red,thick},
        tick label style={anchor=east}},
        extra y tick labels={}]
	\addplot[color=blue] file{images/image_data/louvain_cluster_errors_option4.dat};
	\end{axis}
	
	\begin{axis}[
	    name=young_r,
	    at={($(immune_l.east)+(1cm,3.5cm)$)},
	    anchor=west,
        title=\textsc{young} spectral,
        title style={yshift=-0.2cm},
        width=0.5\columnwidth,
        grid=both,
        grid style={line width=.1pt, draw=gray!20},
        ymax=1,
        height=3.7cm,
		xlabel=$k$,
		ytick distance=0.2,
		xlabel shift=-0.15cm,
		extra y ticks={0.11059757236227824},
        extra tick style={grid=major,major grid style={red,thick},
        tick label style={anchor=east}},
        extra y tick labels={}]
	\addplot[color=blue] file{images/image_data/spectral_cluster_errors_option4.dat};
	\end{axis}
\end{tikzpicture}
    \caption{Error rate $E$ for different parameters per clustering method (resolution parameter $r$ for Louvain, $k$ clusters for spectral) for each corpus in Phase 2.}
    \label{fig:clustering}
\end{figure}
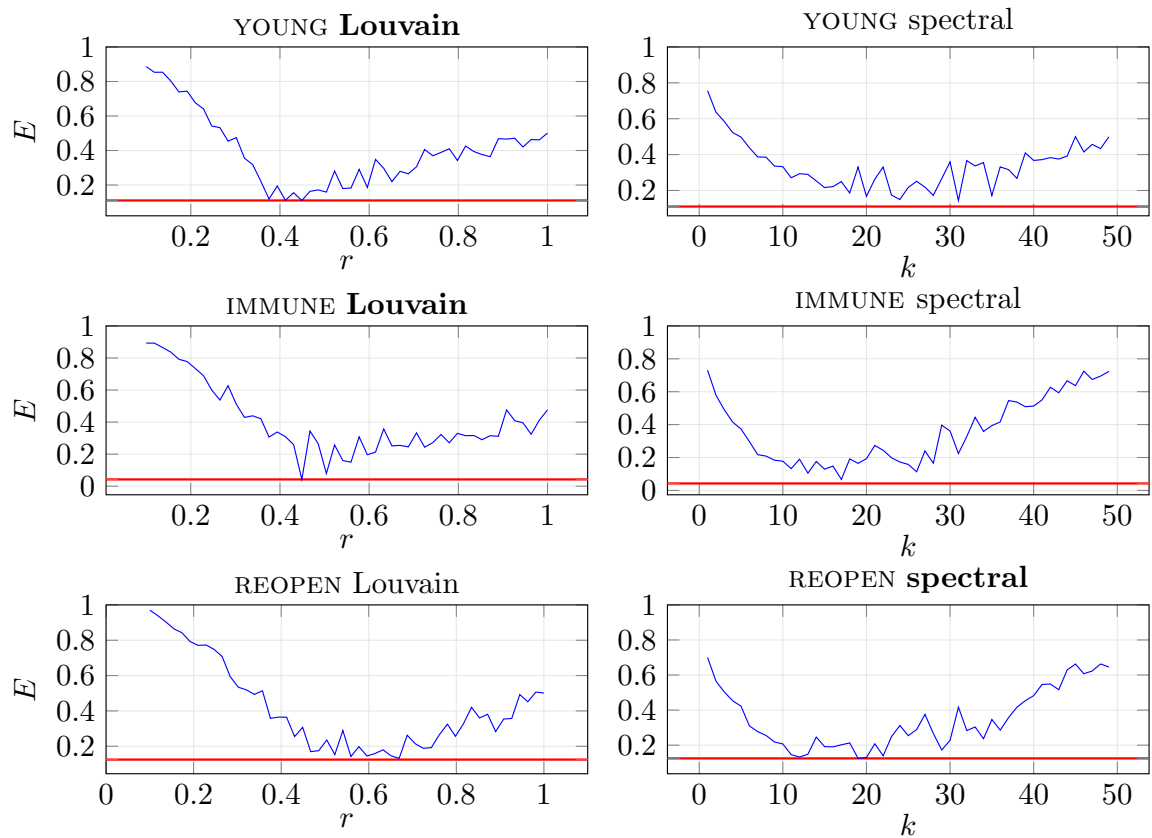

Figure~\ref{fig:clustering} compares Louvain and spectral clustering for extracting argument clusters. Generally, both methods show a clear minimum for obtaining the final argument clusters.
Louvain clustering yields the smallest error for the \young and \immune corpora, and spectral clustering for \reopen corpus. These methods create 20, 14, and 18 clusters respectively. We pick these clusters as input to the argument selection phase.

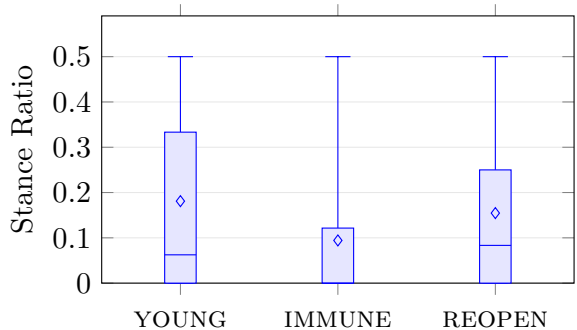
\begin{figure}[tb]
    \centering
    \begin{tikzpicture}
    \begin{axis} [ybar=2pt,
        name=plot-stance,
        boxplot/draw direction=y,
        width=0.5\columnwidth,
        height=5cm,
        xtick = data,
        xmin = 1,
        xmax = 3,
        ymin = 0,
        ymax = 0.59,
        bar width=17pt,
        ymajorgrids,
        grid style={line width=.1pt, draw=gray!20},
        xtick={1,2,3},
        xticklabels={\textsc{young}, \textsc{immune}, \textsc{reopen}},
        x label style={at={(axis description cs:0.5,-0.4)},anchor=north},
        x tick label style={yshift=-0.2em},
        ylabel={Stance Ratio},
        legend pos = north west,
        legend style={at={(0.20,-.5)}, anchor=west, /tikz/every even column/.append style={column sep=0.5cm}},
        legend columns=1,
        /pgfplots/boxplot/box extend=0.2,
		ytick distance=0.1,
        legend image code/.code={%
          \draw[#1] (0cm,-0.1cm) rectangle (0.6cm,0.1cm);
        },
        enlarge x limits = {0.25},
    ]
    \addplot+ [
    boxplot prepared={
        upper quartile=0.33333333333333337,
        lower quartile=0.0,
        upper whisker=0.5,
        lower whisker=0.0,
        median=0.0625,
        average=0.18101190476190476,
        draw position=1
    },
    fill={blue!10},
    draw={blue},
    ] coordinates {};
    \addplot+ [
    boxplot prepared={
        upper quartile=0.12152777777777779,
        lower quartile=0.0,
        upper whisker=0.5,
        lower whisker=0.0,
        median=0.0,
        average=0.09424603174603176,
        draw position=2
    },
    fill={blue!10},
    draw={blue},
    ] coordinates {};
    \addplot+ [
    boxplot prepared={
        upper quartile=0.25,
        lower quartile=0.0,
        upper whisker=0.5,
        lower whisker=0.0,
        median=0.08333333333333331,
        average=0.15462962962962962,
        draw position=3
    },
    fill={blue!10},
    draw={blue},
    ] coordinates {};
    \end{axis}
\end{tikzpicture}
    \caption{Stance distribution for clusters extracted for each corpus in Phase 2. A ratio of 0.5 denotes an equal number of pro and con arguments inside a cluster.}
    \label{fig:stance_error}
\end{figure}

Not all arguments inside the same cluster are constrained to have the same stance (pro or con) towards the policy option. We count what proportion of arguments in the cluster do not adhere to the majority stance. The distribution of stances scores is visualized in  Figure~\ref{fig:stance_error}. While we see that the upper limit is that half the arguments in each cluster are not agreeing with the majority label, the average ratio denotes that only a small fraction of argument stances do not agree with the majority stance label. This shows that the clusters generally represent a coherent distribution of arguments with similar stances to each policy option. The ratio on average is lowest for \immune, which is the option with the highest ratio of con opinions. 

\subsection{Phase 3: Key Argument Selection}
In Phase 3, we compare five Argument Selection methods for extracting a representative argument for each of the clusters obtained in Phase 2. We first perform an odd-one-out task to evaluate the coherence of the clusters according to each tested Argument Selection method (see Section~\ref{sec:method-phase-3} for additional details). Then, we evaluate the quality of the arguments that are selected to represent clusters.

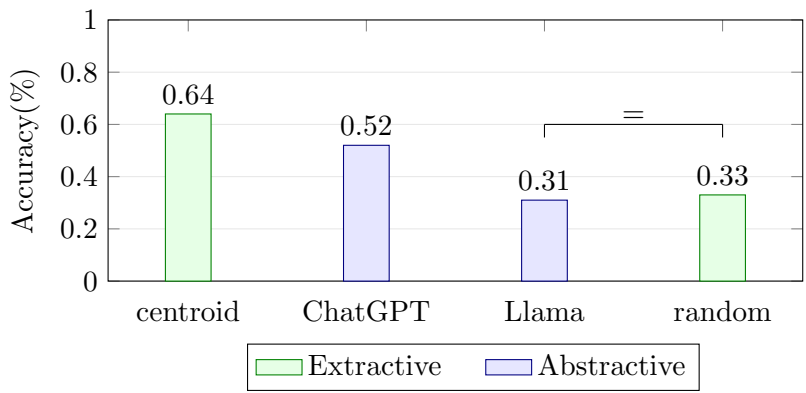
\begin{figure}[tb]
    \centering
        

\begin{tikzpicture}
    \begin{axis}[
        xmin = 1,
        xmax = 4,
        xtick = data,
        x label style={at={(axis description cs:0.5,-0.4)},anchor=north},
        xticklabel style={rotate=0, anchor=north, yshift=-0.3em},
        xtick={1,2,3,4},
        enlarge x limits = {0.15},
        xticklabels={centroid, ChatGPT, Llama, random},
        ymin = 0.0,
        ymax=1.0,
        width=0.7\columnwidth,
        height=5cm,
        legend style={at={(0.20,-.33)}, anchor=west, /tikz/every even column/.append style={column sep=0.5cm}},
        legend columns=2,
        legend image code/.code={%
          \draw[#1] (0cm,-0.1cm) rectangle (0.6cm,0.1cm);
        },
        ylabel=Accuracy(\%),
        nodes near coords,
        nodes near coords align={vertical},
        ymajorgrids,
        grid style={line width=.1pt, draw=gray!20},
        bar width=17pt,
        ]
        \addplot[ybar,fill={green!10},draw={green!50!black}] coordinates {(1, 0.64)};
        \addplot[ybar,fill={blue!10},draw={blue!50!black}] coordinates {(2, 0.52)};
        \addplot[ybar,fill={blue!10},draw={blue!50!black}] coordinates {(3, 0.31)};
        \addplot[ybar,fill={green!10},draw={green!50!black}] coordinates {(4, 0.33)};
    
    \draw[black] (axis cs:3,0.6) -- (axis cs:4,0.6);
    \draw[black] (axis cs:3,0.55) -- (axis cs:3,0.6);
    \draw[black] (axis cs:4,0.55) -- (axis cs:4,0.6);
    
    
    
    
    \node at (axis cs:3.5,0.63) (star1) {=};

    \legend{Extractive, Abstractive, , };
    
    \end{axis}
\end{tikzpicture}
    \caption{Accuracy on the odd-one-out task per method. Key Argument Selection methods marked with $=$ do not significantly differ ($p<0.05$) in their error proportions.}
    \label{fig:selection-odd-one}
\end{figure}

\paragraph{Odd-one-out task}
Figure~\ref{fig:selection-odd-one} shows the results of the odd-one-out evaluation. We perform pairwise statistical analysis by employing McNemar’s test \shortcite{dietterich1998approximate} with Holm-Bonferroni correction on multiple tests \shortcite{aickin1996adjusting}. The test results indicate whether methods significantly differ in their misclassifications. We observe that only \emph{Llama--random} does not have a significant difference in error proportions and can thus be assumed to perform similarly to each other. Conversely, two out of three methods outperform the random baseline. This indicates that these methods identify cluster membership relatively consistently with the results of \hyena, although with considerable error rates. For Llama, we encountered a strong position bias with respect to the ordering of the triple: independently of which was the odd-one-out argument, the model primarily picks arguments at a specific index. This causes its performance to be similar to random picking. We attribute this to the lack of instruction tuning for the Llama model.

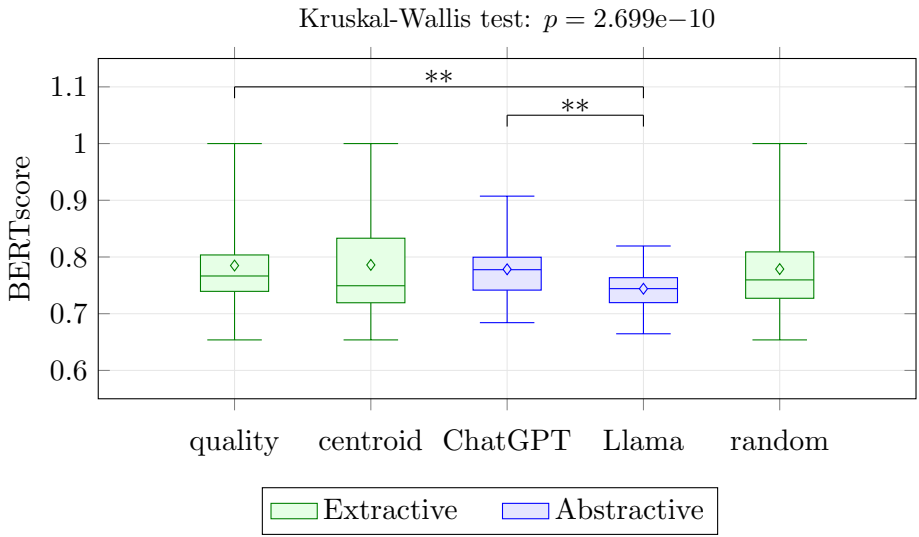
\begin{figure}[tb]
    \centering
    \begin{tikzpicture}
    \begin{axis} [ybar=4pt,
        name=plot-bertscore,
        boxplot/draw direction=y,
        width=0.8\columnwidth,
        title={\small Kruskal-Wallis test: $p=2.699\mathrm{e}{-10}$},
        height=6cm,
        xtick = data,
        grid=both,
        grid style={line width=.1pt, draw=gray!20},
        xmin = 1,
        xmax = 5,
        ymin = 0.55,
        ymax = 1.15,
        xtick={1,2,3,4,5},
        xticklabels={quality, centroid, ChatGPT, Llama, random},
        x label style={at={(axis description cs:0.5,-0.4)},anchor=north},
        xticklabel style={rotate=0, anchor=north, yshift=-0.3em},
        ylabel={BERTscore},
        legend pos = north west,
        legend style={at={(0.20,-.33)}, anchor=west, /tikz/every even column/.append style={column sep=0.5cm}},
        legend columns=2,
        /pgfplots/boxplot/box extend=0.5,
		ytick distance=0.1,
        legend image code/.code={%
          \draw[#1] (0cm,-0.1cm) rectangle (0.6cm,0.1cm);
        },
        enlarge x limits = {0.25},
    ]
    \addplot+ [
    boxplot prepared={
    upper quartile=0.8036,
    lower quartile=0.7394,
    upper whisker=1.0000,
    lower whisker=0.6538,
    median=0.7665,
    average=0.7849,
        draw position=1
    },
    fill={green!10},
    draw={green!50!black},
    ] coordinates {};
    \addplot+ [
    boxplot prepared={
        upper quartile=0.8331,
        lower quartile=0.7193,
        upper whisker=1.0000,
        lower whisker=0.6538,
        median=0.7493,
        average=0.7860,
        draw position=2
    },
    fill={green!10},
    draw={green!50!black},
    ] coordinates {};
    \addplot+ [
    boxplot prepared={
        upper quartile=0.7996,
        lower quartile=0.7416,
        upper whisker=0.9073,
        lower whisker=0.6842,
        median=0.7776,
        average=0.7783,
        draw position=3
    },
    fill={blue!10},
    draw={blue},
    ] coordinates {};
        \addplot+ [
    boxplot prepared={
        upper quartile=0.7635,
        lower quartile=0.7195,
        upper whisker=0.8194,
        lower whisker=0.6646,
        median=0.7442,
        average=0.7440,
        draw position=4
    },
    fill={blue!10},
    draw={blue},
    ] coordinates {};
        \addplot+ [
    boxplot prepared={
        upper quartile=0.8090,
        lower quartile=0.7272,
        upper whisker=1.0000,
        lower whisker=0.6538,
        median=0.7596,
        average=0.7787,
        draw position=5
    },
    fill={green!10},
    draw={green!50!black},
    ] coordinates {};
    \legend{Extractive, , ,Abstractive};
    \draw[black] (axis cs:1,1.1) -- (axis cs:4,1.1);
    \draw[black] (axis cs:1,1.08) -- (axis cs:1,1.1);
    \draw[black] (axis cs:4,1.08) -- (axis cs:4,1.1);
    \node at (axis cs:2.5,1.11) (star1) {**};
    \draw[black] (axis cs:3,1.05) -- (axis cs:4,1.05);
    \draw[black] (axis cs:3,1.03) -- (axis cs:3,1.05);
    \draw[black] (axis cs:4,1.03) -- (axis cs:4,1.05);
    \node at (axis cs:3.5,1.06) (star1) {**};
    \end{axis}
\end{tikzpicture}
    \caption{Aggregated BERTScore for the different Key Argument Selection methods across all corpora and argument clusters (Phase 3). Method pairs indicated by ** differ significantly from each other in median performance ($p<0.05$). }
    \label{fig:bertscore-arg-selection}     
\end{figure}

\paragraph{Evaluating Argument Selection}
To select the best-performing Argument Selection method, we compare BERTScores in Figure~\ref{fig:bertscore-arg-selection}. We use the Kruskal-Wallis test (a non-parametric alternative to ANOVA since the scores are not normally distributed) to test whether all medians are equal at a 5\% significance level \shortcite{kruskal1952use}. Since we obtain a score well below our threshold, we conduct a post-hoc follow-up to identify pairs of significantly different Key Argument Selection methods. We employ Dunn's multiple comparisons of mean rank sums \shortcite{dunn1964multiple} with Holm-Bonferroni correction on multiple tests \shortcite{aickin1996adjusting}. 

All extractive methods have a higher standard deviation than the generative methods. Some selected representative arguments likely caused the high maxima for extractive methods, since they are copied verbatim from opinions in the corpus. Conversely, the low minima are due to the extractive methods' inability to find representatives from the cluster (since there may be noisy clusters, see Figure~\ref{fig:selection-odd-one}). 
For the abstractive methods, the lower bound is higher, showing how rephrasing the selected argument makes it more related to all arguments inside a cluster. Between the abstractive methods, ChatGPT has a higher standard deviation than Llama. Since we did not perform extensive prompt engineering, there is room for improvement in both methods with better-crafted prompts.

The only significantly different method is Llama, with all others achieving similar BERTScore performance. Surprisingly, none of the approaches on average performs considerably better than random. This suggests that selecting a representative argument from the cluster is relatively simple in practice because the argument clusters are sufficiently coherent. However, in the final evaluation, humans will be judging the match between selected arguments and individual opinions. Here, we strive for a better worst-case performance---we care less about having perfect matches, but rather wish to have fewer misrepresentation errors. Thus, given the comparable averages, we opt for the method with the highest lower boundary (the abstractive methods) and higher median score (ChatGPT outperforms Llama significantly), which we use for the remainder of the experiments.

Finally, we compare the output of Phase 3 of \hyena against a version where the selection was manual. In particular, we take the extractions from Phase 1 and re-evaluate them using a new set of annotators. In Table~\ref{tab:compare-arg-sel}, we show the difference in Precision (Equation~\ref{eq:precision-same-hyena}).

We find that the addition of Argument Selection on average has a slight negative impact on the ability of annotators to match opinions and arguments. Most interestingly, when comparing argument matches for the same set of opinions before and after the addition of Argument Selection, we find that there is only fair agreement between the re-matched labels (Cohen $\kappa=0.255$). This indicates that the argument selection phase makes annotating the match for some opinions to selected key arguments easier while making others more difficult. Selecting arguments using ChatGPT generates key arguments that are representative of the entire cluster, which can be more general than the arguments extracted by annotators from individual opinions. On the one hand, this can cause external annotators to not recognize the specific argument from a given opinion. On the other hand, it may result in annotators matching opinions and arguments on a more abstract level.
\begin{table}[tb]
    \centering
    \begin{tabular}{
            @{}       
    >{\arraybackslash}p{\dimexpr 0.24\linewidth-\tabcolsep}
    >{\centering\arraybackslash}p{\dimexpr 0.19\linewidth-2\tabcolsep}
    >{\centering\arraybackslash}p{\dimexpr 0.19\linewidth-2\tabcolsep}
    >{\centering\arraybackslash}p{\dimexpr 0.19\linewidth-2\tabcolsep}
    >{\centering\arraybackslash}p{\dimexpr 0.19\linewidth-\tabcolsep}
    @{} 
    }
          \toprule
          \textbf{Method} & \young & \immune & \reopen & \textbf{Overall}\\
          \midrule
          \hyena & 0.816 & 0.833  & 0.641 & 0.765\\
          \hyena w/o Phase 3 & 0.787  & 0.848 & 0.739 & 0.789\\
         \bottomrule
    \end{tabular}
    \caption{Comparing Precision ($P$) scores with and without Phase 3 (Key Argument Selection phase). }
    \label{tab:compare-arg-sel}
\end{table}

\subsection{Comparison with Automated Baseline}
\label{sec:results-comparison}
\begin{figure}[tb]
    \centering
    \small
    \begin{tikzpicture}
\def\barx{17} 
\def\barspace{10} 
\def\figwidth{0.7} 

    \begin{axis} [ybar=0pt,
        name=plot-coverage,
	    anchor=west,
        width=\figwidth\columnwidth,
        height=3.4cm,
        bar width = \barx pt,
        xtick = data,
        grid=both,
        grid style={line width=.1pt, draw=gray!20},
        ymin = 0,
        ymax = 1,
        compat = 1.3,
        ytick align=outside,
        xticklabels={\textsc{young}, \textsc{immune}, \textsc{reopen}},
        xticklabel style={rotate=0, anchor=north,xshift=0em, yshift=0.3em},
        xlabel={Coverage},
        x label style={at={(axis description cs:0.5,1.33)},anchor=north},
        legend cell align=left,
        legend style={
            at={(1,0.5)},
            draw=none,
            anchor=west,
            /tikz/every even column/.append style={
                column sep=0.25cm
            }
        },
        legend columns=1,
        legend image code/.code={%
          \draw[#1] (0cm,-0.1cm) rectangle (0.3cm,0.1cm);
        },
        enlarge x limits = {.25},
    ]
        \addplot+[draw = red!70!gray!70,
            line width = 0.1mm,
            fill = red!70!gray!10,
            bar shift=-\barspace pt,
        ] coordinates {(1, 0.3478) (2, 0.4785) (3, 0.3289)};
        \addplot [draw = red,
            line width = 0.1mm,
            pattern = crosshatch dots,
            pattern color = red,
            bar shift=-\barspace pt,
        ] coordinates {(1, 0.1765) (2, 0.1333) (3, 0.1647)};
        \addplot [draw = blue!50,
            line width = 0.1mm,
            fill = blue!50,
            bar shift=\barspace pt,
        ]   coordinates {(1, 0.4118) (2, 0.2706) (3, 0.3490)};
        \legend {$C_A$, $C_A^{common}$, $C_H = C_H^{common}$};
    \end{axis}
    \begin{axis} [ybar=0pt,
        name=plot-precision,
	    at={($(plot-coverage.east)-(0cm,3.7cm)$)},
        anchor=east,
        width=\figwidth\columnwidth,
        height=3.4cm,
        bar width = \barx pt,
        xtick = data,
        grid=both,
        grid style={line width=.1pt, draw=gray!20},
        ymin = 0,
        ymax = 1,
        compat = 1.3,
        ylabel near ticks,
        ytick align=outside,
        xticklabels={\textsc{young}, \textsc{immune}, \textsc{reopen}},
        xticklabel style={rotate=0, anchor=north,xshift=0em, yshift=0.3em},
        xlabel={Precision},
        x label style={at={(axis description cs:0.5,1.33)},anchor=north},
        legend cell align=left,
        legend style={
            at={(1,0.5)},
            draw=none,
            anchor=west,
            /tikz/every even column/.append style={
                column sep=0.25cm
            }
        },
        legend columns=1,
        legend image code/.code={%
          \draw[#1] (0cm,-0.1cm) rectangle (0.3cm,0.1cm);
        },
        enlarge x limits = {0.25},
    ]
        \addplot [draw = red,
            line width = 0.1mm,
            bar shift=-\barspace pt,
            pattern = crosshatch dots,
            pattern color = red,            
        ]   coordinates {(1, 0.6667) (2, 0.5294) (3, 0.5)};
        
        \addplot [draw = blue!50,
            line width = 0.1mm,
            bar shift=\barspace pt,
            fill= blue!50,
        ]   coordinates {(1, 0.816) (2, 0.833) (3, 0.641)};
        \legend {$P_A^{common}$, $P_H^{common}$};

    \end{axis}

    \begin{axis} [ybar=0pt,
        name=plot-diversity,
	    at={($(plot-precision.east)-(0cm,3.7cm)$)},
        anchor=east,
        width=\figwidth\columnwidth,
        height=3.4cm,
        bar width = \barx pt,
        xtick = data,
        grid=both,
        grid style={line width=.1pt, draw=gray!20},
        ymin = 0,
        ymax = 1,
        compat = 1.3,
        ylabel near ticks,
        ytick align=outside,
        xticklabels={\textsc{young}, \textsc{immune}, \textsc{reopen}},
        xticklabel style={rotate=0, anchor=north,xshift=0em, yshift=0.3em},
        xlabel={Diversity},
        x label style={at={(axis description cs:0.5,1.33)},anchor=north},
        legend cell align=left,
        legend style={
            at={(1,0.5)},
            draw=none,
            anchor=west,
            /tikz/every even column/.append style={
                column sep=0.25cm
            }
        },
        legend columns=1,
        legend image code/.code={%
          \draw[#1] (0cm,-0.1cm) rectangle (0.3cm,0.1cm);
        },
        enlarge x limits = {0.25},
    ]
        \addplot [draw = red,
            line width = 0.1mm,
            bar shift=-\barspace pt,
            pattern = crosshatch dots,
            pattern color = red,            
        ]   coordinates {(1, 0.2444) (2, 0.3235) (3, 0.2381)};
        \addplot [draw = blue!50,
            line width = 0.1mm,
            bar shift=\barspace pt,
            fill= blue!50,
        ]   coordinates {(1, 0.5512) (2, 0.5859) (3, 0.4957)};
        
        \legend {$D_A$, $D_H$};

    \end{axis}

\end{tikzpicture}
    \caption{Comparing \hyena and ArgKP.}
    \label{fig:coverage-eval}
\end{figure}
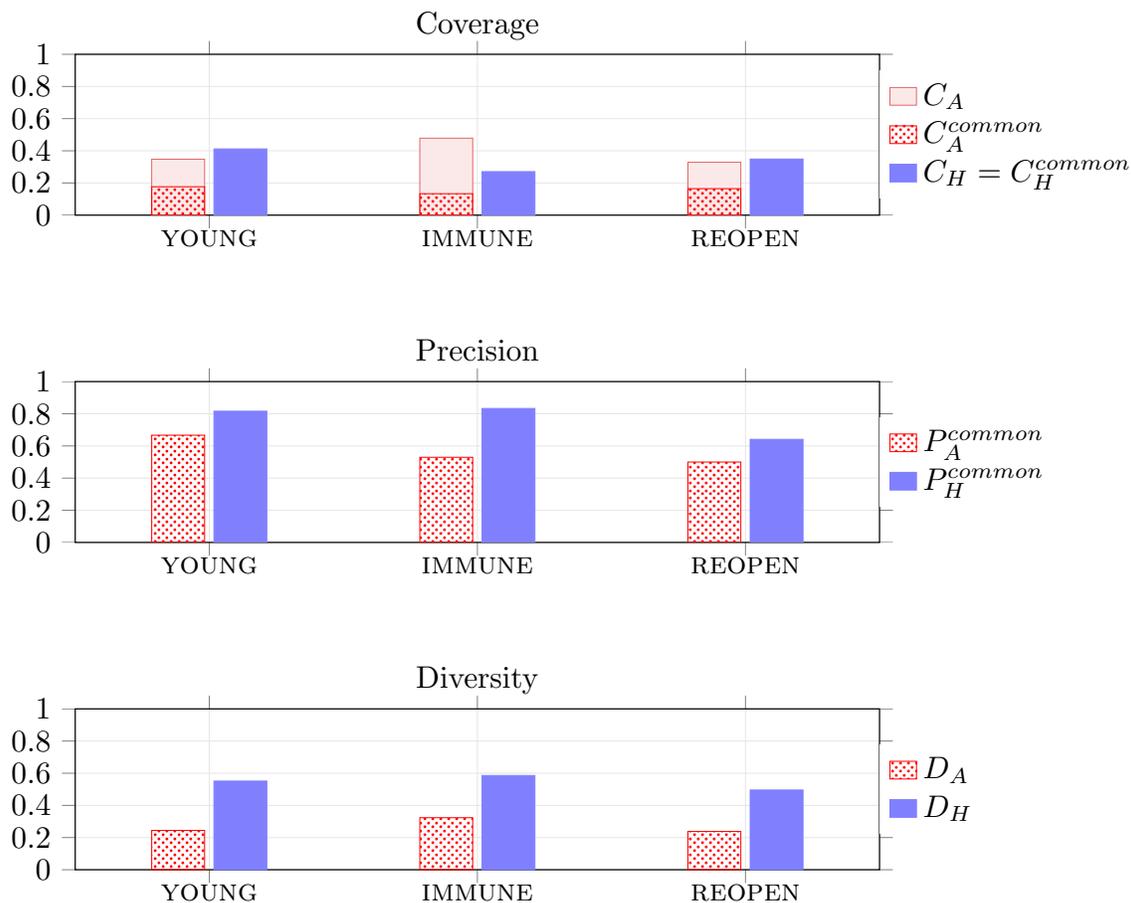

Figure~\ref{fig:coverage-eval} compares the coverage, precision, and diversity scores of \hyena and \argkp. The low coverage (for both methods) indicates that a large number of opinions do not map to a key argument. This is not surprising since real-world opinions are noisy. 

Considering \emph{all} observed opinions ($C_H$ and $C_A$), \hyena yields slightly higher coverage than \argkp in the \young and \reopen corpora. In contrast, \argkp yields higher coverage than \hyena in the \immune corpus. We attribute this to the repeated arguments in the \immune corpus. As 83\% of opinions are con-opinions, the \immune policy option (Table~\ref{tab:data-example}) was highly opposed and its corpus contains many repeated arguments. Since the set of \emph{all} observed opinions is the entire corpus for \argkp, the repeated arguments inflate its coverage. However, since \hyena is designed to observe only a small subset of diverse opinions from the corpus, the repeated arguments do not influence its coverage significantly. This is corroborated in the diversity scores, where we observe \hyena to consistently output a set of arguments that is more diverse than the ones produced by \argkp. 

In addition to comparing coverage over \emph{observed} opinions, we compare the coverage of \hyena and \argkp with respect to a \emph{common} set of diverse opinions. In this comparison ($C_H^{common}$ and $C_A^{common}$), \hyena yields consistently higher coverage (0.34 on average) than \argkp (0.16 on average) in all three corpora. \argkp often fails to recognize the key arguments in the diverse set of opinions included by \hyena.

\argkp yields a larger number of key arguments (around 30 for each option) than \hyena. However, these arguments lead to an average precision of 0.56. In contrast, \hyena extracts fewer argument clusters (on average 17 per option), but with higher precision (0.80). 

\subsection{Comparison with Manual Baseline}
Table~\ref{tab:confusion} shows counts of overlapping (yes, yes), missing (no, yes), and new (yes, no) key arguments between \hyena and the manual baseline. 
\hyena required an analysis of 765 opinions, compared to the estimated 3,000 opinions seen in the manual baseline. Despite the lower human effort, the \hyena lists largely overlap with the expert lists.

\begin{table}[tb]
    \centering
    \begin{tabular}{
                @{}       
    >{\arraybackslash}p{\dimexpr 0.125\linewidth-\tabcolsep}
    >{\centering\arraybackslash}p{\dimexpr 0.125\linewidth-2\tabcolsep}
    >{\centering\arraybackslash}p{\dimexpr 0.125\linewidth-2\tabcolsep}
    >{\centering\arraybackslash}p{\dimexpr 0.125\linewidth-2\tabcolsep}
    >{\centering\arraybackslash}p{\dimexpr 0.125\linewidth-2\tabcolsep}
    >{\centering\arraybackslash}p{\dimexpr 0.125\linewidth-2\tabcolsep}
    >{\centering\arraybackslash}p{\dimexpr 0.125\linewidth-2\tabcolsep}
    >{\centering\arraybackslash}p{\dimexpr 0.125\linewidth-\tabcolsep}
    @{}}
        \toprule
        & & &  \multicolumn{5}{c}{\textbf{Manual baseline}} \\
        \cmidrule(lr){3-8}
        & & \multicolumn{2}{c}{\young} & \multicolumn{2}{c}{\immune} & \multicolumn{2}{c}{\reopen}\\
        \cmidrule(lr){3-4}\cmidrule(lr){5-6} \cmidrule(lr){7-8}
         & &  yes & no & yes & no & yes & no \\
        \midrule
        \multirow{2}{*}{\textbf{\hyena}} & yes &  8 & 7 &  7 & 2 & 10 & 1 \\
        & no    & 1 & -- & 0 & -- & 4 & --\\
        \bottomrule
    \end{tabular}
    \caption{A confusion matrix comparing the key argument lists generated by \hyena and manual baseline.}
    \label{tab:confusion}
    
\end{table}

\hyena missed some key arguments that the experts identified, e.g., a key argument about building herd immunity was not in the \hyena list for the \reopen option. 
We conjecture that increasing the number of opinions annotated in \hyena would subsequently yield the missing insights. 
\hyena also led to new insights that experts missed, e.g., an argument about the physical well-being of young people was not on the expert list for the \young option. 
Likely, the larger (random) sample of opinions experts analyzed did not include opinions supporting this argument, whereas the smaller (intelligently selected) set sampled in \hyena did.

\section{Discussion}
We find that \hyena exploits the strengths of automated methods and the insights from human annotation. \hyena outperformed an automated KPA model in terms of precision and diversity, and on a diverse set of opinions, can capture more nuanced arguments. Further, \hyena expanded beyond an expert analysis, showing how a fully manual procedure may also be limited. In the remainder of this section, we expand on three specific aspects.

\label{sec:discussion}
\paragraph{Limitations} Our experimental setup and comparisons are limited in their scope in multiple ways, thus making our conclusions hard to generalize. Our choice of baseline is the ArgKP model, which was optimized for the task of extracting Key Arguments from a corpus of opinions. However, other automated baselines are conceivable, especially with the introduction of the current generation of flexible LLMs (e.g., ChatGPT, Llama). Those models may be employed for KPA by using prompting techniques \shortcite{liu2023pre}. The capabilities of these models seem to imply that they have access to higher order argumentation knowledge \shortcite{lauscher2022scientia}, and thus would fare better than the basic ArgKP model. However, having such LLMs reliably process large amounts of citizen feedback without hallucinations is a nontrivial task, and the danger of models synthesizing ungrounded arguments exists \shortcite{halluc}. In this process, due diligence to preserve a variety of perspectives is required (e.g., by optimizing for a range of opinions instead of single-annotator labels, \shortciteR{bakker2022fine,vandermeer2024annotatorcentric}) in order to prevent rampant misrepresentation of marginalized demographics.

Instead of relying solely on the judgment of an LLM for the task of KPA, we opted to include one in the final step of \hyena. While some of the criticism for using an LLM for end-to-end KPA still holds for the Argument Selection step as well, our method investigated a more controlled setup, supported by an objective task definition. Through our comparisons with random and human-generated labels, we aim to show where, how, and to what extent LLMs may aid in the KPA process. As ever, the choice of metrics remains important for measuring the effect size.

\paragraph{Balancing Task Allocation} The pairwise comparison in the consolidation phase is the most human-intensive task in \hyena, and the effort increases with the number of analyzed opinions. Also, comparing arguments is cognitively demanding, partly evidenced by the low IRR. While \hyena reduced the number of comparisons required in the consolidation phase by 60\%, we may experiment with different setups or other techniques for comparing arguments to remove this overhead. For example, first clustering the key arguments and then consolidating the arguments within these clusters (reverse order as \hyena) may drastically reduce the number of judgments required in the second phase. Furthermore, future versions of HyEnA could benefit from investigating why annotators disagreed on labels in each phase, as it can lead to possible improvements in the annotation task.

We place human efforts in places where there are multiple bidirectional benefits possible stemming from performing the task. For instance, the Argument Annotation task both serves the purpose of analyzing the opinions to progress our method, as well as actively making annotators perform \emph{perspective-taking}. On multiple occasions, annotators noted their increase in sympathy and recognition of the issues raised in the comments, showcasing how the task could further help bring understanding to a group of citizens.

\paragraph{Ablations studies} All parts of the \hyena pipeline are open to adjustment and can be performed by humans, machines, or a combination. In this work, we presented a specific version of this pipeline, but other ways of combining humans and AI are possible. However, the impact of choosing specific components remains unclear for parts of the pipeline, since we experimented with a single algorithm in some cases (e.g., the use of \textsc{Power} in Key Argument Consolidation, or the LLMs in Key Argument Selection). 

\hyena presents a general framework that allows individual phases to be supported by different types of technologies and different groups of crowd/expert annotators. Within this hybrid framework, we considered the following criteria when deciding to allocate tasks to humans or AI methods:
\begin{enumerate*}[label=(\arabic*)]
    \item let humans read other's opinions to promote perspective-taking,
    \item use humans to solve tasks where AI methods may incur considerable error,
    \item leverage AI methods for routine tasks, and
    \item use task-specific intrinsic evaluation metrics for selecting the right method. 
\end{enumerate*}

In each phase, we perform both intrinsic evaluation (e.g., observe error rates for particular tasks or annotator behavior) and extrinsic evaluation against two baselines. This fits a standardized machine learning pipeline, except that we are now able to \begin{enumerate*}[label=(\arabic*)]
    \item evaluate annotator behavior and model performance jointly, and
    \item make decisions on which techniques to use based on some intermediate statistic. 
\end{enumerate*} We believe this setup to be generalizable for Hybrid Intelligence systems, as it makes the role of the designer and their decisions explicit \shortcite{akata2020research}. Furthermore, the results remain interpretable, as any decision made by either annotators or models can be traced from opinion to selected key argument.

Different configurations of the \hyena framework are possible, and the one we have presented is an instance that tackles the problem of policy feedback analysis. \hyena is a complex combination of AI methods and human annotation. Our main objective was to present the \hyena framework, as well as a real-world use case to show the benefit of using a Hybrid Intelligent methodology. However, other choices for individual components of \hyena can be used, or parts of the method can be performed solely by humans or AI methods. We leave this open for future work, as swapping out components is not straightforward and requires considerable amounts of work. We envision research to come up with similar use cases where HI can make a significant impact. 

\section{Conclusion and Future Directions}
\label{sec:conclusion}
We develop and evaluate \hyena, a hybrid method that combines human judgments with automated methods to generate a diverse set of key arguments. 
\hyena extracts key arguments from noisy opinions and achieves consistent coverage, whereas the coverage of a state-of-the-art automated method drops by 50\% when switching from all (containing repeated) opinions to diverse opinions.
Moreover, the key arguments extracted by \hyena are more precise than those extracted by the automated baseline. 
Additionally, \hyena provides important insights that were not included in an expert-driven analysis of the same corpus, despite requiring fewer opinions to be analyzed.

Finding arguments in a discourse is only one aspect that constitutes the perspectives in a discussion. Future work can incorporate analysis of other perspective factors, such as values \shortcite{liscio2022values,van-der-meer-etal-2023-differences}, sentiment, emotion, and attribution \shortcite{van2016grasp}. By combining these rich aspects with arguments, we can merge the logical basis of the discussion with other semantic and syntactic information, allowing close scrutiny of the perspectives in opinions.

\paragraph{Acknowledgements}
This research was (partially) funded by the Hybrid Intelligence Center, a 10-year program funded by the Dutch Ministry of Education, Culture, and Science through the Netherlands Organisation for Scientific Research (NWO). 
We would like to thank the anonymous reviewers whose insightful comments and suggestions helped improve the paper. 

\bibliography{final_papers,extra}
\bibliographystyle{theapa}

\end{document}